\documentclass[runningheads]{llncs}
\usepackage{graphicx}

\usepackage{tikz}
\usepackage{comment}
\usepackage{amsmath,amssymb} 
\usepackage{color}

\usepackage{verbatim}
\usepackage{mdwlist}
\usepackage{mdframed}
\usepackage{booktabs}
\usepackage{color,soul}
\usepackage[ruled,vlined]{algorithm2e}

\usepackage[pagebackref=true,breaklinks=true,letterpaper=true,colorlinks,bookmarks=false]{hyperref}

\newcommand{\squishlist}{
 \begin{list}{$\bullet$}
  { \setlength{\itemsep}{0pt}
     \setlength{\parsep}{3pt}
     \setlength{\topsep}{3pt}
     \setlength{\partopsep}{0pt}
     \setlength{\leftmargin}{1.5em}
     \setlength{\labelwidth}{1em}
     \setlength{\labelsep}{0.5em} } }

\newcommand{\squishend}{
  \end{list}  }

\newcommand{\beginsupplement}{%
        \setcounter{table}{0}
        \renewcommand{\thetable}{S\arabic{table}}%
        \setcounter{figure}{0}
        \renewcommand{\thefigure}{S\arabic{figure}}%
     }

\definecolor{lightorange}{rgb}{1,0.98,0.95}
\definecolor{darkred}{rgb}{0.5,0,0}

\begin{document}
\pagestyle{headings}
\mainmatter
\def\ECCVSubNumber{1319}  

\title{Explainable Face Recognition} 

\titlerunning{Explainable Face Recognition}
%
\author{
Jonathan R. Williford\inst{1}\orcidID{0000-0002-9178-2647} \and
Brandon B. May\inst{1}\orcidID{0000-0002-9914-2441} \and
Jeffrey Byrne\inst{1,2}\orcidID{0000-0001-8973-0322}}
%
\authorrunning{J.R. Williford et al.}
%
\institute{
Systems \& Technology Research, Woburn, MA 01801, USA\\
\url{https://www.stresearch.com}\\
\email{\{jonathan.williford,brandon.may\}@stresearch.com}\\
\and
Visym Labs, Cambridge, MA 02140, USA\\
\email{jeff@visym.com}}
\maketitle

\begin{abstract} 
Explainable face recognition (XFR) is the problem of explaining the matches returned by a facial matcher, in order to provide insight into why a probe was matched with one identity over another. In this paper, we provide the first comprehensive benchmark and baseline evaluation for XFR.  We define a new evaluation protocol called the ``inpainting game'', which is a curated set of 3648 triplets (probe, mate, nonmate) of 95 subjects, which differ by synthetically inpainting a chosen facial characteristic like the nose, eyebrows or mouth creating an inpainted nonmate. An XFR algorithm is tasked with generating a network attention map which best explains which regions in a probe image match with a mated image, and not with an inpainted nonmate for each triplet. This provides ground truth for quantifying what image regions contribute to face matching. Finally, we provide a comprehensive benchmark on this dataset comparing five state-of-the-art XFR algorithms on three facial matchers.  This benchmark includes two new algorithms called subtree EBP and Density-based Input Sampling for Explanation (DISE) which outperform the state-of-the-art XFR by a wide margin.
\end{abstract}

\section{Introduction}

Explainable AI \cite{19explainable} is the problem of interpreting, understanding and visualizing machine learning models.  Deep convolutional network trained at large scales are traditionally considered blackbox systems, where designers have an understanding of the dataset and loss functions for training, but limited understanding of the learned model.  Furthermore, predictions generated by the system are often not explainable as to why the system generated this output for that input.  An explainable AI system would enable interpretation of what the ML model has learned  \cite{Ribeiro16}\cite{Bau2017NetworkDQ}, enable transparency to understand and identify biases or failure modes in the system \cite{pmlr-v81-buolamwini18a}\cite{nist19}\cite{Georgetown18}\cite{Raji2019ActionableAI} and provide user friendly visualizations to build user trust in critical applications \cite{Selvaraju2016}\cite{Simonyan2014}\cite{zhou2015cnnlocalization}.  

Explainable face recognition (XFR) is the problem of explaining why a face matching system matches faces.  Human adjudicators have a long history in explaining face recognition in the field of forensic face matching.  Professional facial analysts follow the FISWG standards \cite{FISWG12} which leverage comparing facial morphology, measuring facial landmarks and matching scars, marks and blemishes.   These features are used to match a controlled mugshot of a proposed candidate  to an uncontrolled probe, such as a security camera image.  However, these approaches require a candidate list for human adjudication, and a candidate list in a modern workflow is returned from a facial matching system \cite{Phillips2018FaceRA}.  Why did the face matching system return that candidate list for this probe?  What facial features did the face matching system use, and are they the same as the FISWG standards?  Is the face matcher biased or noisy?  The goal of XFR is to explore such questions, and answer why a system matched a pair of faces.  A successful explainable system would increase confidence in a face matching system for professional examiners, enable intepretation of the internal face representations by machine learning researchers and generate trust by the user community.  

What is an ``explanation'' in face recognition?  Explainable AI has explored various forms of explanation for machine learning systems in the form of: activation maximization \cite{Simonyan2013DeepIC}, synthesizing optimal images \cite{Nguyen2016SynthesizingTP}, network attention \cite{Zhang2016}\cite{petsiukrise}\cite{Fong2017}, network dissection \cite{Bau2017NetworkDQ} or synthesizing linguistic explanations \cite{Hu2018ExplainableNC}.  However, a key challenge in explainable AI is the lack of ground truth to compare and quantify explainable results across networks. XFR is especially challenging because the difference between near-mates or {\em doppelgangers} is subtle, the explanations are non-obvious, and differences are rarely well localized in a compact facial feature \cite{Castanon18}.  

\begin{figure}[t!]
    \centering
    \includegraphics[width=\linewidth]{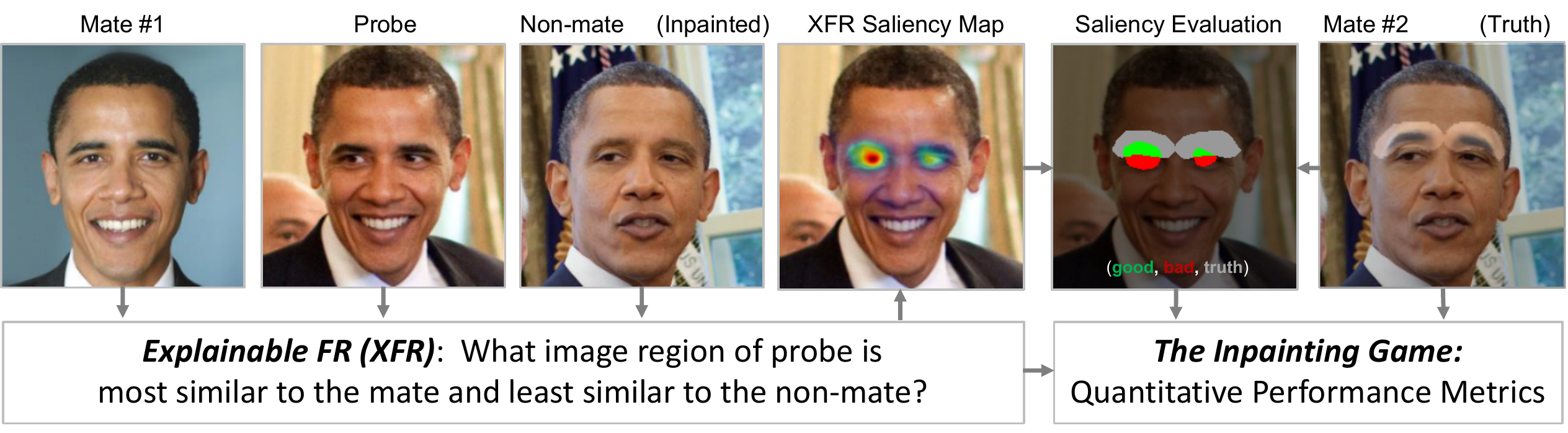} 
    \caption{Explainable Face Recognition (XFR).
    Given a image triplet of ({\it probe}, {\it mate 1}, {\it nonmate}), an explainable face recognition algorithm is tasked with estimating which pixels belong in a region that is discriminative for the mate - i.e. a region is more similar to the mate than the non-mate.
    These estimations are given as a saliency map.
    The nonmate has been synthesized by inpainting a given region (e.g. eyebrows) that changes the identity according to the given network.
    This provides ground truth for a quantitative evaluation of XFR algorithms using the ``inpainting game'' protocol.    
    }
     \label{f:xfr_overview}
\end{figure}

In this paper, we provide the first comprehensive benchmark for explainable face recognition (XFR).  Fig. \ref{f:xfr_overview} shows the structure of this problem.  An XFR system is given a triplet of ({\it probe}, {\it mate}, {\it nonmate}) images.  The XFR system is tasked with generating a saliency map that best captures the regions of the probe image that increase similarity to the mate and decrease similarity to the nonmate.  This provides an explanation for why the matcher provides a high verification score for the pair ({\it probe}, {\it mate}) and a low verification score for ({\it probe}, {\it nonmate}).  This explanation can be quantitatively evaluated by synthesizing nonmates that differ from the mate only in specific regions (e.g. nose, eyes, mouth), such that if the saliency algorithm selects these regions, then it performs well on this metric. 
This paper makes the following contributions:

\begin{enumerate}
    \item {\bf XFR baseline}.  We provide a baseline for XFR based on five algorithms for network attention evaluated on three publicly available convolutional networks trained for face recognition: LightCNN \cite{wu2018light}, VGGFace2 \cite{Cao18} and ResNet-101. These baselines include two new algorithms for network attention called subtree EBP (Sec. \ref{s:weighted-subtree-ebp}) and DISE (Sec. \ref{s:dise}).  
    \item {\bf Inpainting game protocol and dataset}.  We provide a standardized evaluation protocol (Sec. \ref{s:inpainting_game}) and dataset (Sec. \ref{s:dataset}) for fine grained discriminative visualization of faces.  This provides a quantitative metric for objectively comparing XFR systems.
    \item {\bf XFR evaluation}.  We provide the first comprehensive evaluation of XFR using the baseline algorithms on the inpainting game protocol to provide a benchmark for future research (Sec. \ref{s:inpainting_game_results}).  Furthermore, in the supplemental material, we show a qualitative evaluation on novel (non inpainted) images to draw conclusions about the utility of the methods for explanation on real images. 
\end{enumerate}

 \section{Related Work}
 



The related work most relevant for our proposed approach to XFR can be broadly categorized into two areas:  network attention models for convolutional networks and interpretable face recognition.  

Network attention is the problem of generating an image based saliency map which visualizes the input regions that best explains a class activation output of a network.  Gradient-based methods \cite{Selvaraju2016}\cite{Simonyan2014}\cite{zhou2015cnnlocalization} attempt to compute the the derivative of the class signal with respect to the input image,    
while other approaches \cite{cao2015look} modify network architectures to capture these signals or localize attribution \cite{Kindermans2017}.  Excitation backprop \cite{Zhang2016}, contrastive EBP \cite{Zhang2016} or truncated contrastive EBP \cite{Castanon18} formulate the saliency map as marginal probabilities in a probabilistic absorbing Markov chain.  Layerwise relevance propagation \cite{Li2017BeyondSU}\cite{Wojciech2016}\cite{Bach2015} provides network attention through a set of layerwise propagation rules theoretically justified by deep Taylor decomposition.
Latent attention networks learn an auxiliary network to map input to attention, rather than exploring the network directly \cite{Grimm2017}.  
Inversion methods \cite{mahendran15understanding} seek to recover natural images that have the same feature representation as a given image. However, the same insights have not yet been applied to fine grained categorization for face recognition.  Finally, black box methods have explored network attention for systems that do not have an exposed convolutional network \cite{Dabkowski2017}\cite{petsiukrise}\cite{Fong2017}\cite{cao2015look}.  The approaches to XFR explored in this paper are most closely related to EBP \cite{Zhang2016}, RISE \cite{petsiukrise}, and methods for network attention for pairwise similarity \cite{Stylianou2019}.

Recently, there has been emerging research on the interpretation 
of face recognition systems\cite{towards-interpretable-face-recognition-2019}\cite{Zee_2019_ICCV}\cite{Castanon18}\cite{Xu2018DeeperIO}\cite{RichardWebesterFace2018}\cite{Dhar19}\cite{Zhong19}.  Visual psychophysics \cite{RichardWebesterFace2018} have provided a set of tools for the controlled manipulation of input stimuli and metrics for the output responses
evoked in a face matching system. This approach was inspired by Cambridge Face Memory Test \cite{Duchaine2006TheCF}, which involves progressively perturbing face images using a chosen transformation function (e.g. adding noise) 
to investigate controlled degradation of matching performance \cite{RichardWebesterFace2018}.  This approach enables detailed studies of the failure modes of a face matcher or exploring how facial attributes are expressed in a network \cite{Dhar19}\cite{Zhong19}.  In contrast, our approach generates controlled degradations using inpainting, to provide localized ground truth for evaluation of network attention models.  
In \cite{towards-interpretable-face-recognition-2019}, the authors propose a novel loss function to encourage part separability during network dissection of parts in a convolutional network for face matching.  This approach is primarily concerned with training new networks to maximize interpretability, rather than studying existing networks.  In \cite{Zee_2019_ICCV}, the authors study pairwise matching of faces, to visualize features that lead towards classification decisions.  This is similar in spirit to our proposed approach, however we provide a performance metric for evaluating a saliency approach as well as extending visualizations to mated and nonmated triplets.  Finally, in \cite{Xu2018DeeperIO}, the authors visualize the features of shape and texture that underlie subject identity decisions.  This approach uses 3D modeling to generate a controlled dataset, rather than inpainting.  However, given the authors conclusions that texture has a much larger effect of matching than morphology, having a ground truth dataset that includes texture variation would be an appropriate metric for explainable face recognition.

\section{Explainable Face Recognition (XFR)}
\label{s:xfr} 

XFR is the problem of explaining why a face matcher matches faces.  Fig. \ref{f:xfr_overview} shows the structure of this problem.  Given a triplet of ({\it probe}, {\it mate}, {\it nonmate}), the XFR algorithm is tasked with generating a saliency map that explains the regions of the probe image that maximize the similarity to the mate and minimize the similarity to the nonmate.  This provides an explanation for why the matcher returns this image for the mated identity.  

Why triplets?  Previous work has shown that pairwise similarity between faces is heavily dominated by the periocular region and nose \cite{Castanon18}, as confirmed by the qualitative visualization study performed in 
the supplementary material
The periocular region and nose is almost always used for facial classification, but this level of XFR is not very helpful in explaining finer levels of discrimination.
Our goal is to highlight those regions for a probe that are more similar to a presumptive mate and {\em simultaneously} less similar to a nonmate.  This triplet of ({\it probe}, {\it mate}, {\it nonmate}) provides a deeper explanation beyond facial class activation maps for the {\em relative importance} of facial regions.

In this section, we describe five approaches for network attention in XFR.  These approaches are all whitebox methods, which assume access to the underlying convolutional network used for facial matching.  The objective of XFR is to generate a non-negative saliency map, that captures the underlying image regions of the probe that are most similar to the mate and least similar to the nonmate.  The XFR algorithm can use any property of the convolutional network to generate this saliency map.  For our benchmark evalution, we selected three state-of-the-art approaches for network attention (excitation backprop, contrastive excitation backprop and truncated contrastive excitation backprop) following the survey and evaluation results in \cite{Castanon18}.  In this section, we introduce two new methods to improve upon these published approaches: subtree EBP (Sec. \ref{s:weighted-subtree-ebp}) and DISE (Sec. \ref{s:dise}).

\begin{figure*}[t]
    \centering
    \includegraphics[width=\linewidth]{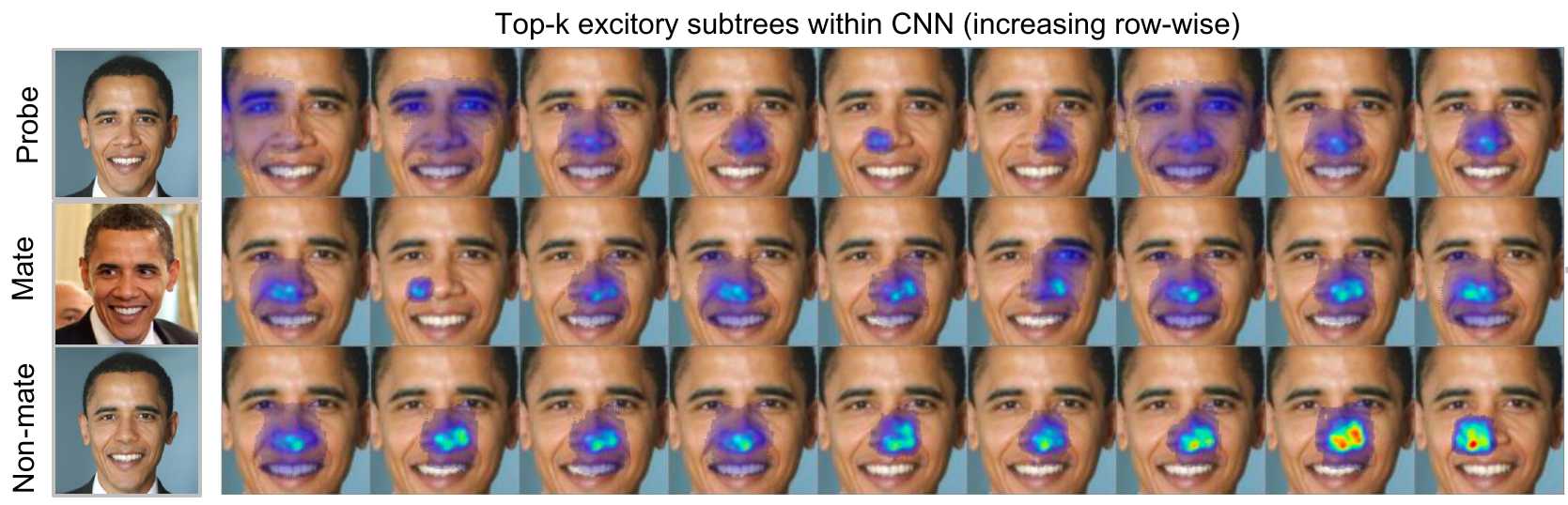}
    \caption{Subtree EBP.  Given a triplet of ({\it probe}, {\it mate}, {\it nonmate}) subtree EBP explores the activations of individual nodes in a convolutional network that minimizes a triplet loss, which maximizes similarity to the mate and minimizes similarity to the nonmate.  The excitory regions for each node are visualized independently, sorted by loss and combined into a saliency map that best explains how to discriminate the probe.}
     \label{f:xfr_whitebox_overview}
\end{figure*}

\subsubsection{Excitation Backprop (EBP).}
\label{s:ebp}
Excitation backprop (EBP) \cite{Zhang2016} models network attention as a probabilistic winner-take-all (WTA) process.
EBP calculates the probability of traversing to a given node in the convolutional network, with the probabilities being defined by the positive weights and non-negative activations.  The output of EBP is a saliency map that localizes regions in the image that are excitory for a given class.  
%
%

In our approach, we replace the cross-entropy loss for EBP with a triplet loss \cite{Schroff15}.  The original formulation of EBP considers a cross-entropy loss to optimize softmax classification of a set of clases in the training set.  In this new formulation, given three embeddings for a mate $(m)$, nonmate $(n)$ and probe $(p)$, the triplet loss function is a max-margin hinge loss
\begin{equation}
\label{e:triplet_loss}
L(p,n,m) = \mathtt{max}(0,~||p-m||^2 - ||p-n||^2 + \alpha).
\end{equation}
This uses the squared Euclidean distance between embeddings to capture similarity, such that the loss is minimized when the distance from the probe and mate is small (similarity is high) and the distance form the probe to the nonmate is large (similarity is low), with margin term $\alpha$.  This loss function extends EBP to cases where a new subject is observed at test time that was not present in the training set, as is commonly the case with face matching systems.  

\subsubsection{Contrastive EBP (cEBP).}
\label{s:contrastive-triplet-ebp}
Contrastive EBP was introduced \cite{Zhang2016} to handle fine-grained network attention for closely related classes.  This approach discards activations common to a pair of classes, to provide network attention specific to one class and not another.  In our approach, contrastive EBP \cite{Zhang2016} is combined with a triplet loss (eq \ref{e:triplet_loss}).    

\subsubsection{Truncated Contrastive EBP (tcEBP).}
Truncated contrastive EBP was introduced \cite{Castanon18} as an extension of cEBP that considers the contrastive EBP attention map only within the kth percentile of the EBP saliency map.  This addresses an observed instability of cEBP \cite{Castanon18} resulting noisy attention maps for faces.

\subsection{Subtree EBP}
\label{s:weighted-subtree-ebp}
In this section, we introduce {\em Subtree EBP}, a novel method for whitebox XFR.  This approach uses the triplet loss function (eq \ref{e:triplet_loss}), with the following extension.  
Given a triplet ({\it probe}, {\it mate}, {\it nonmate}) images, we compute the gradient of the triplet loss function ($\frac{\partial L}{\partial x_i}$) with respect to every node $x_i$ in the network.  This approach uses the standard triplet-based learning, where the mate and nonmate embeddings are assumed constant and the gradient is computed relative to the probe image. Next, we sort the gradients at every node $x_j$ in decreasing order, and select the top-k nodes with the largest positive gradients.  These are the top-k nodes in the network that most affect the triplet loss, to increase the similarity to the mate and simultaneously decrease the similarity to the nonmate.  Finally, we construct $k$ EBP saliency maps $S_i$ starting from each of the selected interior nodes, then the $S_i$ are combined in a weighted convex combination with weights $w_i=\frac{\partial L}{\partial x_i}$ and 
\begin{equation}
\label{e:subtree_ebp}
S=\frac{1}{\sum_j w_j} \sum_i \frac{\partial L}{\partial x_i} S_i    
\end{equation}
where the weights are given by the loss gradient ($w_i$), normalized to sum to one.  This forms the final subtree EBP saliency map $S$.  

Fig. \ref{f:xfr_whitebox_overview} shows an example of the subtree EBP method.  This montage shows the top 27 nodes with the largest triplet loss gradient for the shown triplet.  Each node results in a saliency map corresponding to the excitory subtree rooted at this node.   The weight of the saliency map is proportional to the gradient sorted rowwise, so that the nodes in the bottom right affect the loss more than the nodes in the upper left.  Each of these saliency maps are combined into a convex combination (eq. \ref{e:subtree_ebp}) forming the final network attention map.  In this example, the nonmate was synthesized to differ with the mate only in the nose region, and our method is able to correctly localize this region.  
The supplementary material
shows a more detailed example of this selection process starting from the largest excitation node at each layer of a ResNet-101 network.  This result shows that nodes selected close to the image will be well localized, nodes in the middle of the network correspond to parts and nodes selected close to the embedding correspond to the whole nose and eyes of the face.  

\begin{figure*}[t]
    \centering
    \includegraphics[width=\linewidth]{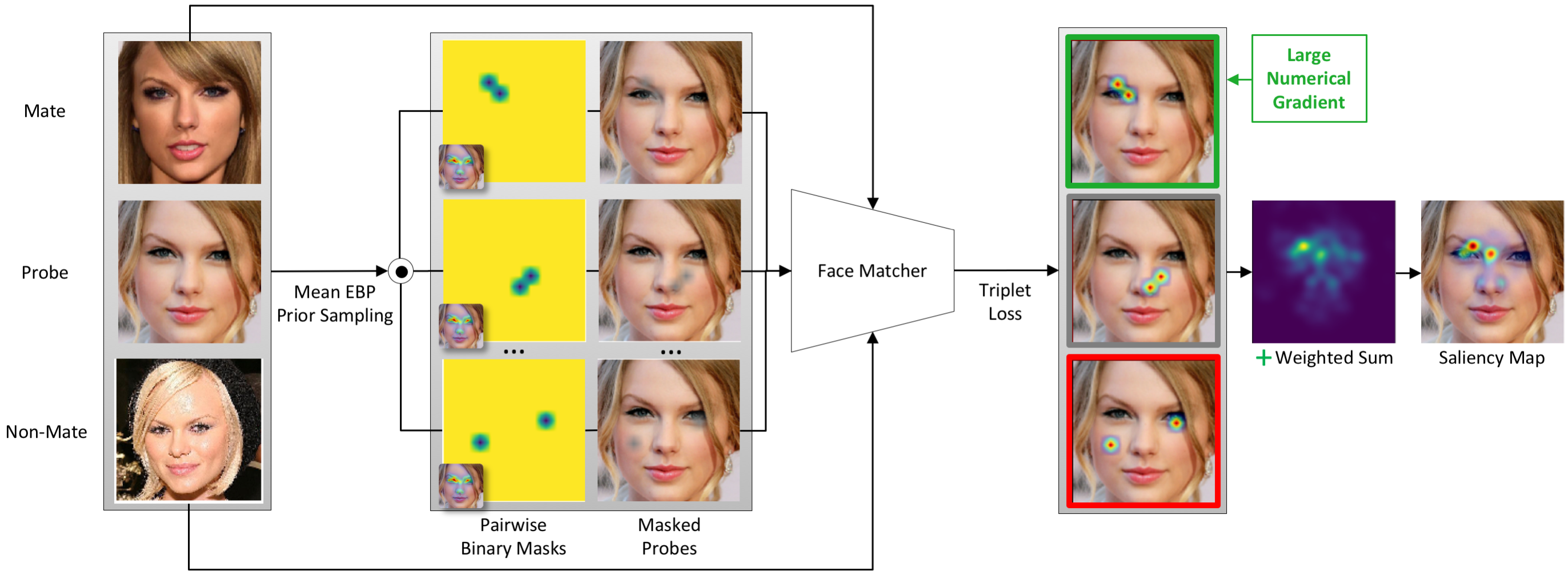}
    \caption{Density-based Input Sampling for Explanation (DISE).  Our approach is an extention of RISE \cite{petsiukrise} for XFR.  This approach occludes small regions in the probe image with grey (i.e. masked pixels), sampled according to a prior density derived from excitation backprop, and computes a numerical gradient for the triplet loss for ({\it probe}, {\it mate}, {\it nonmate}) given these masked probes.  Masks with a large numerical gradient are more heavily weighted in the accumulated saliency map.
    }
     \label{f:bbox_cts_block_diagram}
\end{figure*}

\subsection{Density-based Input Sampling for Explanation (DISE)}
\label{s:dise}

Density-based Input Sampling for Explanation (DISE) is a second novel approach for whitebox XFR introduced in this paper.  DISE is an extension of Randomized Input Sampling for Explanation (RISE) \cite{petsiukrise} using a prior density to aid in sampling. 
Previous work \cite{petsiukrise}\cite{Fong2017} has constructed a saliency map associated with a particular class by randomly perturbing the input image by masking selected pixels, evaluating it using a blackbox system, and accumulating those perturbations based on how confident the system is that the modified input image corresponds to the target class.  However, these approaches generate masks to occlude the input image uniformly at random.  This sampling process is inefficient, and can be improved by introducing a prior distribution to guide the sampling.  In this section, we describe the extension to RISE \cite{petsiukrise} where the prior density for input sampling is derived from a whitebox EBP with triplet loss.  

Fig. \ref{f:bbox_cts_block_diagram} shows an overview of this approach.  Our approach extends RISE \cite{petsiukrise} for XFR as follows:

\begin{enumerate}
    \item Using a non-uniform prior for generating the random binary masks
    \item Restricting the masks to use a sparse, fixed number of mask elements
    \item Defining a numerical gradient of the triplet loss to weight each mask
\end{enumerate}

\subsubsection{Non-Uniform Prior.}  Prior research on discriminative features for facial recognition showed that the most important regions of the face were generally located in and around the eyes and nose (Sec. \ref{s:ebp}). Fig. \ref{f:bbox_cts_block_diagram} shows an example of this saliency map computed for a probe image of Taylor Swift using the VGG-16 \cite{Parkhi15} network as the whitebox face matcher. Using this saliency map as our prior probability for generating random masks allows us to sample the space of most salient masks that will affect the loss more efficiently than assuming a uniform probability across the entire image. Further limiting this prior to the upper 50th percentile of the mean EBP effectively eliminates the possibility of masking out unimportant background elements.

\subsubsection{Sparse Masks.}  Next, we restrict the number of masked elements to be sparse.  RISE considered random binary masks covering the entire input image.  In contrast, we use a sparse mask to highlight the affect of a small localized region of the face on the loss.
We used two mask elements per mask, upsampled by a factor of 12 (to avoid pixel level adversarial effects). We found that filling the masks with a blurred version of the image performed quantitatively better on the inpainting game than using grey masks.

\subsubsection{Numerical gradient.}  Finally, given the probe image which has been masked with the sparse mask sampled from the non-uniform prior, we can compute a numerical gradient of the triplet loss.  Let $p$ be an embedding of the probe, $m$ the mated image embedding, $n$ the nonmated image embedding, and $\hat{p}$ the masked probe embedding.  Then, the numerical gradient of the triplet loss (eq \ref{e:triplet_loss}) can be approximated as:
\begin{equation}
\label{e:dise}
    \begin{aligned}
        \frac{\partial L_{dise}}{\partial p} \approx \mathtt{max}(0,~L(p,m,n)-L(\hat{p},m,n))
    \end{aligned}
\end{equation}
\noindent The numerical gradient is an approximation to the true loss gradient computed by perturbing the input by occluding the probe with a pixel mask, and computing the corresponding change in the triplet loss.  In other words, when the probe masks out a region that increases for the similarity between the probe and mate and decreases for the probe and nonmate, the numerical gradient should be large.  This allows for a loss weighted accumulation of these masks into a saliency map.  The final saliency is accumulated following (eq. \ref{e:subtree_ebp}), where saliency maps $S_i$ are the pairwise binary masks, with non-negative gradient weights  (eq. \ref{e:dise}).   

\section{Experimental Protocol}

Recent explainable AI research has focused on class activation maps
\cite{petsiukrise}\cite{Fong2017}\cite{Selvaraju2016} \cite{Simonyan2014}\cite{zhou2015cnnlocalization}\cite{Zhang2016}\cite{cao2015look}, which visualize salient regions used for classification.
For facial recognition, prior work has shown this is almost always the eyes, nose, and upper lip of the face \cite{Castanon18}.  In facial identification, a probe image is given to a face matching system, which returns the top $K$ identities from a gallery.
A natural question is why the matching system picked the top match instead of the second top match (or remaining top K matches).
One way to give an answer to this question is to highlight the region(s) that match a given identity more than the second identity or other identities.
This saliency map should be larger for the regions that contribute the most to the identity and not others.
In this paper, our goal is to highlight the regions that are responsible for matching a given image to one identity versus a similar identity.

A key challenge for evaluating the performance of an XFR approach is generating ground truth.  For XFR, ground truth not only depends on the selection of probes, mates, and nonmates, but can also depend on a target network for evaluation.  
We address this issue by synthesizing inpainted nonmates or {\em doppelgangers}, where a select region of the face is changed from the original identity.  
Only the inpainted region differs between the two images and therefore only the inpainted region can be used to discriminate between them.  Furthermore, we synthesize doppelgangers based on their ability to reduce the match score for a target network.  We call our overall strategy for quantitative evaluation the {\em inpainting game}.  

\subsection{The Inpainting Game}
\label{s:inpainting_game}

\begin{figure*}[t]
    \centering
    \includegraphics[width=1.0\linewidth]{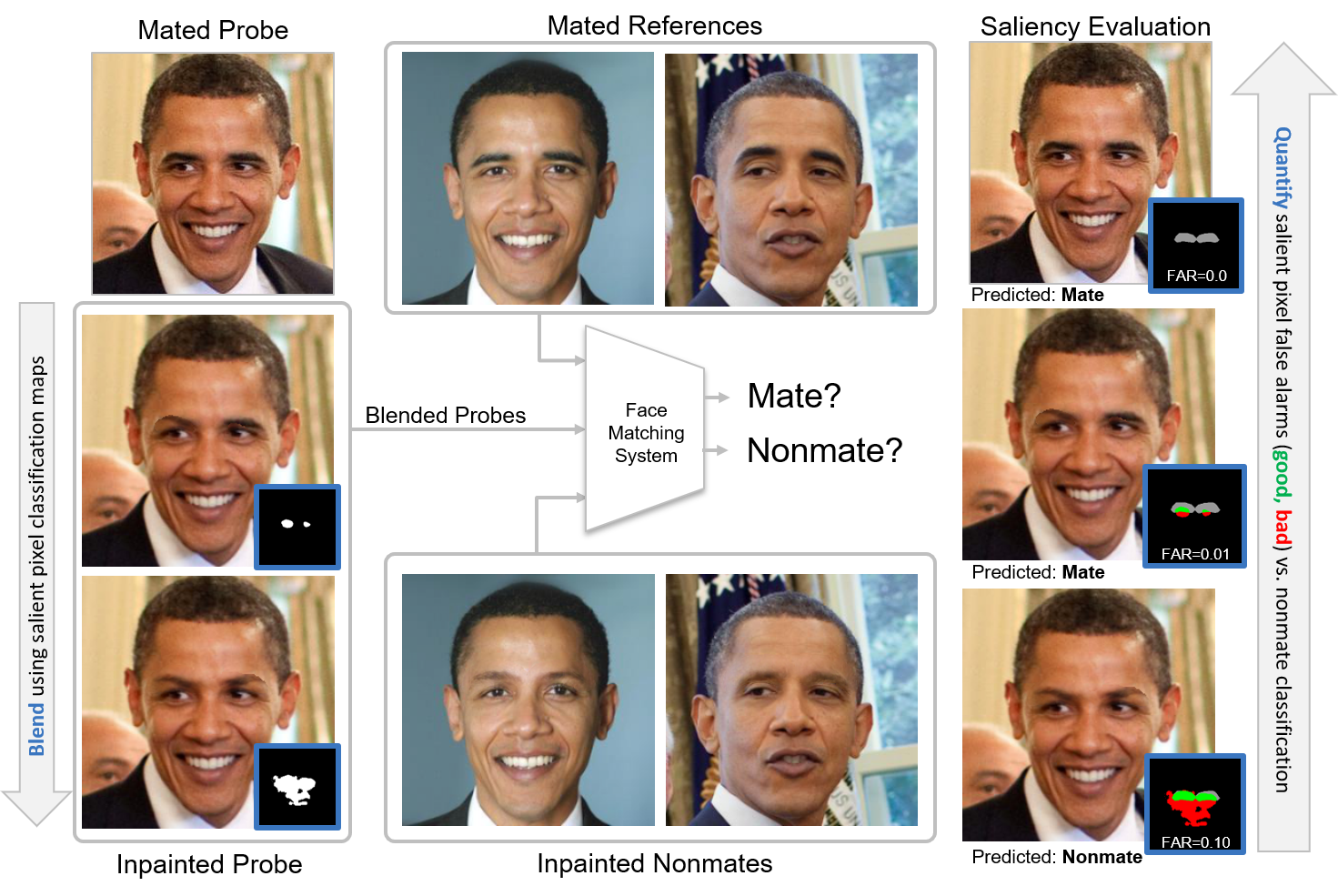}
    \caption{
    Inpainting game overview.  The XFR algorithm is given triplets (probe, mate, nonmate) labeled in the figure as (mated probe, mated references, inpainted non-mates),
    and is tasked with estimating a discriminative saliency map that estimates the likelihood that a pixel belongs to a region that is discriminative for the mate.
    A threshold is applied to the saliency map to classify each pixel as being discriminatively salient (inset, blue squares, left).
    A high performing XFR algorithm will correctly classify the discriminative pixels within the inpainted region (green, right) while avoiding classifying the pixels that are identical between the mated references and the probes as being discriminative (red, right).
   See sec. \ref{s:inpainting_game} for more details. 
    }
     \label{f:inpainting-game-eval}
\end{figure*}

An overview of the inpainting game evaluation is given in Fig. \ref{f:inpainting-game-eval}.
The inpainting game uses four (or more) images for each evaluation:  a probe image, mate image(s), an inpainted probe and inpainted nonmate(s).   The inpainted probe or {\em probe doppelganger} is subtly different from the probe in a fixed region of the face, such as the eyes, nose or mouth.  Similarly, the inpainted nonmate or {\em mate doppelganger} is subtly different from the mate image, such that the doppelgangers are a different identity.  The inpainted probe and inpainted nonmate are constrained to be the same new identity.  Sec. \ref{s:dataset} discusses the construction of this dataset.  


The XFR algorithm is given triplets of probes, mates and nonmates,
labeled in Fig. \ref{f:inpainting-game-eval} as (``mated probe'', ``mated references'' and ``inpainted non-mates'').
For each triplet, the XFR algorithm is tasked with estimating
the likelihood that each pixel belongs to a region that is discriminative for matching the probe to the mated identity over the nonmated/inpainted identity.
These discriminative pixel estimations form a saliency map.
Each pixel is classified as being discriminatively salient by applying a threshold, which forms a binary saliency map.
For each binary saliency map, pixels in the probe are replaced with the pixels from an inpainted probe forming a blended probe.
The inpainted probe is generating by inpainting the same facial region as the inpainted nonmates and is not provided to the XFR algorithm, 
which is sequestered and used for evaluation only.
The saliency map is evaluated by how quickly it can flip the identity of the blended probe from the mate to the non-mate, while maximizing saliency (green) in ground truth (grey) while minimizing false alarms (red).  See Sec. \ref{s:evaluation_methodology} for additional details, including the metrics for the inpainting game. 

\subsection{Inpainting Dataset for Facial Recognition}
\label{s:dataset}

\begin{figure*}[t!]
    \centering
    \includegraphics[width=\linewidth]{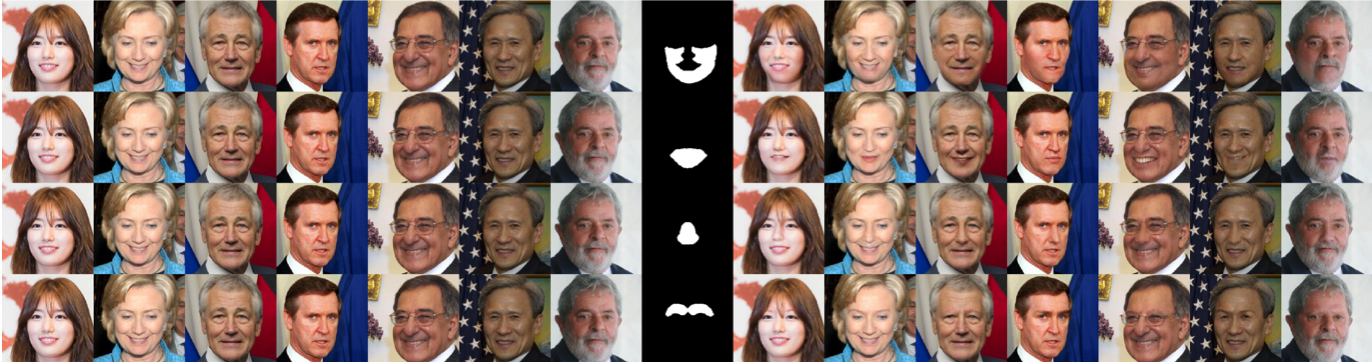}
    \caption{
    Example facial inpainted images.  This montage shows the first four of eight inpainting regions (cheeks, mouth, nose, eyebrows, left face, right face, left eye, right eye), and synthesized inpainted doppelgangers images. The first seven columns show seven subjects, with the same image repeated along the column.  
    The middle column shows a binary inpainting mask that defines the inpainting region.  
    The last seven columns show the inpainted doppelganger image using the mask region for that row, such that the inpainted image differs from the original image only in the mask region.  Observe that the identity may change subtly while looking down a column.
    }
     \label{f:example-probes-inpainted}
\end{figure*}

The inpainting dataset for face recognition is based on the images from the IJB-C dataset \cite{maze2018iarpa}.
The inpainting dataset contains 561 images of 95 subjects selected from IJB-C, for an average of 5.9 images per subject.  We defined eight facial regions for evaluation: 1) cheeks and jaw, 2) mouth, 3) nose, 4) left eye, 5) right eye, 6) eyebrows, and 7) left face, 8) right face.  Each image is inpainted for each of the eight regions forming a total of 4488 inpainted doppelgangers.  From this set, we define 3648 triplets, such that each triplet is a combination of (probe, mate and inpainted doppelganger nonmate).
The XFR algorithms should not be evaluated on triplets for networks that cannot distinguish the original and inpainted identities. Hence, the only the triplets that contain discriminable identities are included for the network the algorithm is being evaluated on.

The inpainted doppelgangers are generated as follows. 
In order to systematically mask the regions, we use the pix2face algorithm \cite{pix2face2017} to fit a 3d face mesh onto each facial image.
We then projected the facial region masks onto the images.
We use pluralistic inpainting \cite{zheng2019pluralistic} to synthesize an image completion in that masked region.
Fig. \ref{f:example-probes-inpainted} shown examples of these inpainted doppelgangers.

A key challenge of constructing the inpainting dataset is to enforce that the inpainted nonmate is in fact a different identity.  Most of our inpainted images are not sufficiently different in similarity from the original mated identity for a specific network.
A given triplet of ({\it probe}, {\it mate}, and {\it inpainted nonmate}) is only included in the dataset if a given target network can distinguish the two identities for the mate/mate doppelganger and the probe/probe doppelganger.
They are required to be able to distinguish these identities both using a nearest match protocol and an verification protocol, such that the verification match threshold for a target network is calibrated to a false alarm rate of 1e-4.
Specifically, each triplet has to fulfill the following criteria in order to be included in the dataset for a given network:
\begin{enumerate}
    \item The original probe must be: (i) more similar to the original/mated identity than the corresponding inpainted/nonmated identity and (ii) 
    correctly verified as the original/mated identity at the calibrated verification threshold.
    \item The inpainted probe must be: (i) more similar to the corresponding inpainted / nonmated identity than the original identity and (ii) correctly verified as the same identity as the inpainted/nonmated identity at the calibrated verification threshold.
\end{enumerate}

The inpainting dataset is filtered for each target network according to the above criteria, resulting in a dataset specific to that target network.  For example, for the ResNet-101 based system, the final filtered dataset includes 84 identities and 543 triplets, filtered down from 95 identities and 3648 triplets.  
Lower performing networks will generally have fewer triplets satisfying the selection criteria than higher performing networks, because they will not be able to discriminate as many of the subtle changes in the inpainted probes.

\subsection{Evaluation Metrics}
\label{s:evaluation_methodology}

The XFR algorithms estimate the likelihood that each pixel belongs to a region that is discriminative for matching the probe to the mated identity over the nonmated/inpainted identity.
These discriminative pixel estimations form a saliency map, where the brightest pixels are estimated to be most likely to belong to the discriminative region.
Fig. \ref{f:inpainting-game-eval} 
shows an example and saliency predictions at two thresholds, where the saliency prediction is shown at different thresholds as a binary mask.

\begin{figure}[tb!]
    \centering
    \includegraphics[width=0.495\linewidth]{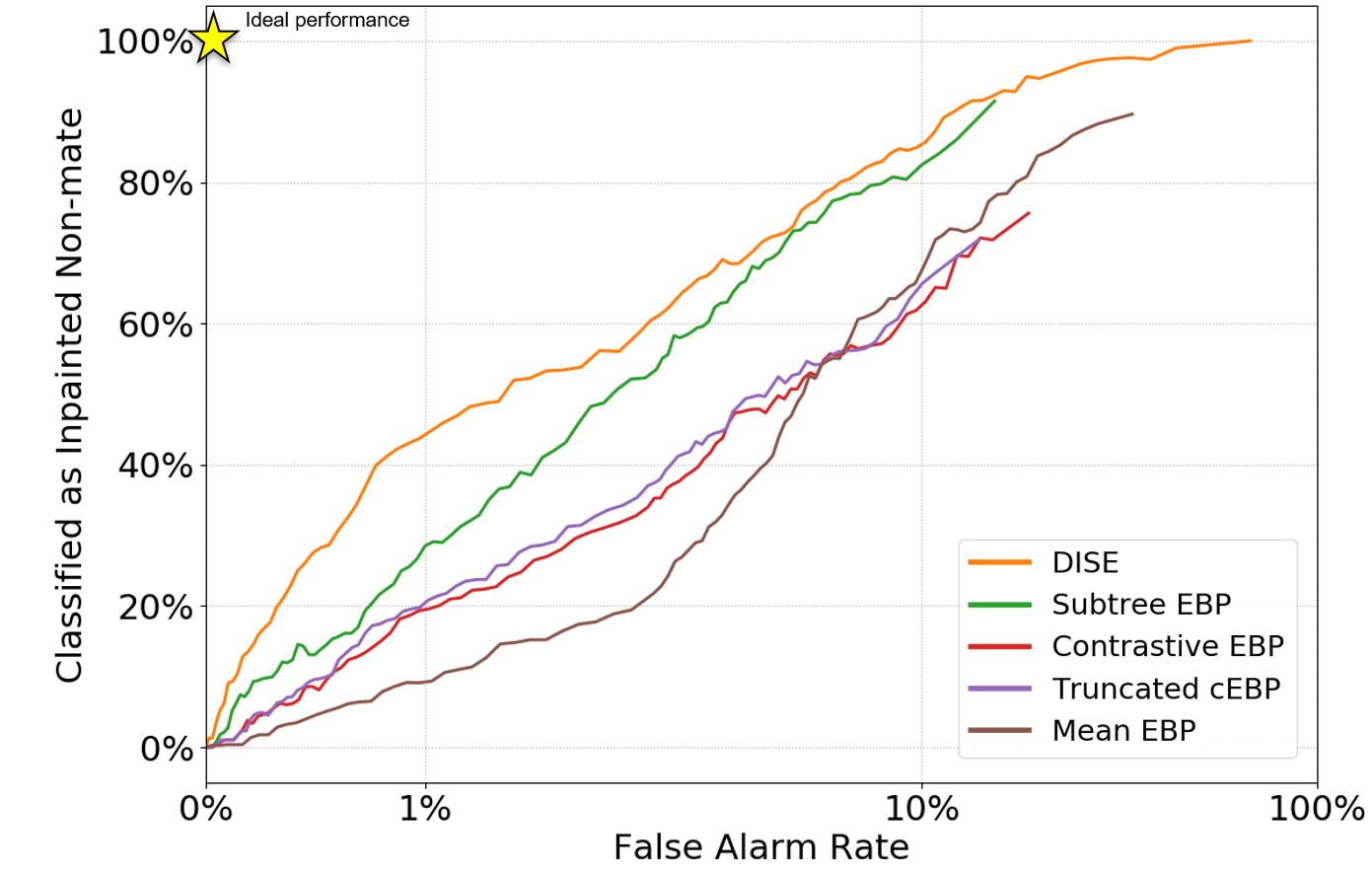}
    \includegraphics[width=0.495\linewidth]{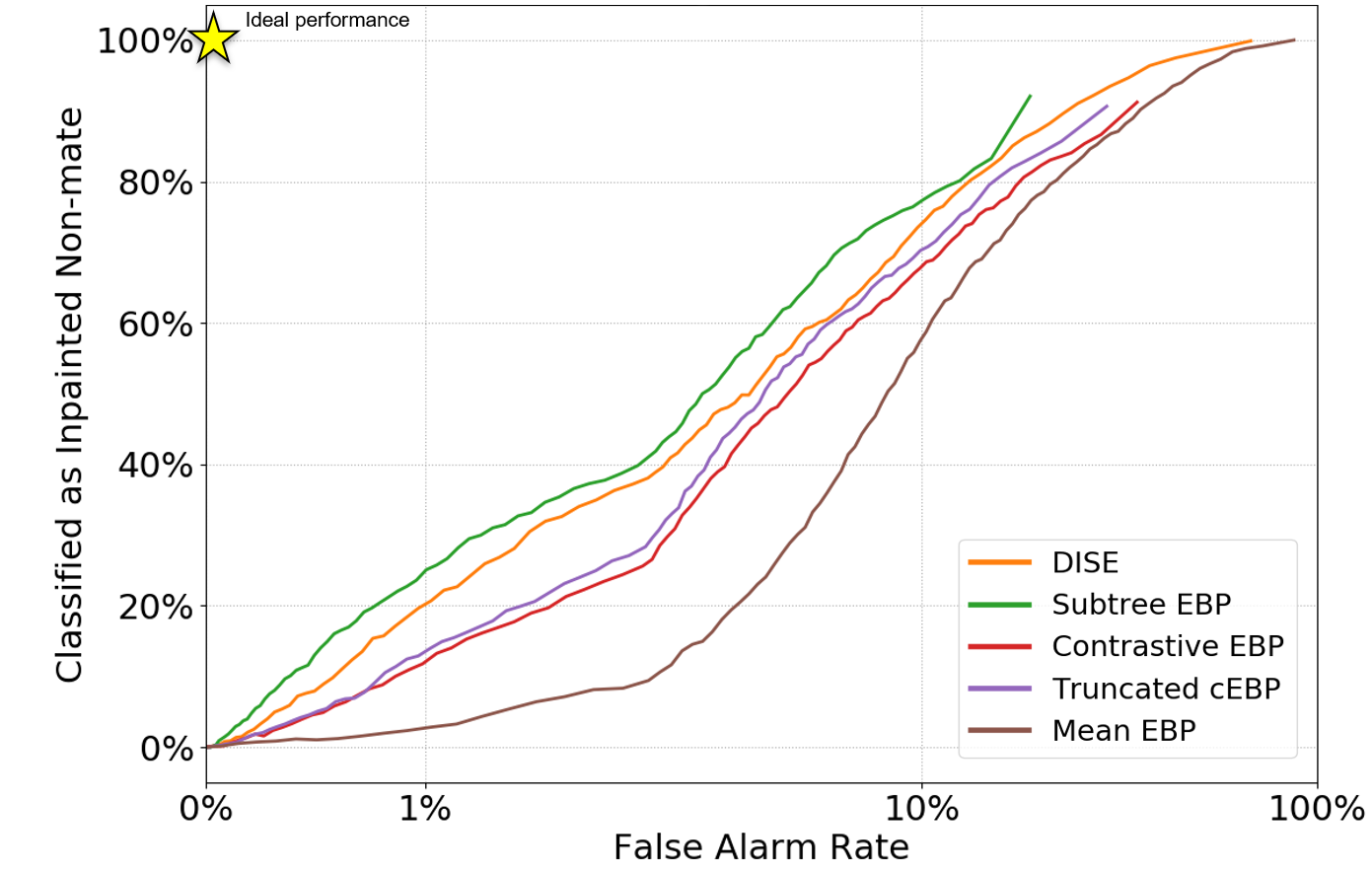}
    \caption{(left) Inpainting game analysis using VGGFace2 ResNet-50.  (right) Inpainting game analysis using Light-CNN.  Refer to Table \ref{t:inpainting_game_results_summary} for a summary of performance at fixed operating points on these curves.
    }
     \label{f:inpainting_game_results_plots}
\end{figure}

In order to motivate our proposed metric, let's first consider using a classic receiver operating characteristic (ROC) curve for evaluation of the inpainting game rather than our proposed metric.  A ROC curve can be generated by sweeping a threshold for the pixel saliency estimations, and computing true accept rate and false alarm rate by using the inpainted region as the positive / salient region and the non-inpainted region as the negative / non-salient region
(i.e. middle column in Fig. \ref{f:example-probes-inpainted}).  However, not all pixels within the inpainted region contribute equally to the identity, and the saliency algorithm should not be either penalized or credited with this selection.  

To address this key challenge, we use mean nonmate classification rate instead of true positive rate for saliency classification.  We play a game where the pixels classified as being salient by sweeping the saliency threshold are replaced with the pixels from the ``inpainted probe'', which is not provided to the saliency algorithm.
These ``blended probes'' can then be classified as original identity or inpainted nonmate identity by the network being tested.
High performing XFR algorithms will correctly assign more saliency for the inpainted regions that will change the identity of the blended probes without increasing the false alarm rate of the pixel salience classification.  This is the key idea behind our evaluation metric.  
The false positive rate is calculated from salient pixel classification across all triplets, using the ground truth masks for the blended probe.
The mean nonmate classification rate is weighted by the number of triplets within each facial region for a filtered dataset, to avoid bias for subprotocols with more examples. 
Example of the output curves for this metric is shown in Fig. 
\ref{f:inpainting_game_results_plots}.

\begin{table}[t]
    \centering
    \includegraphics[width=\linewidth]{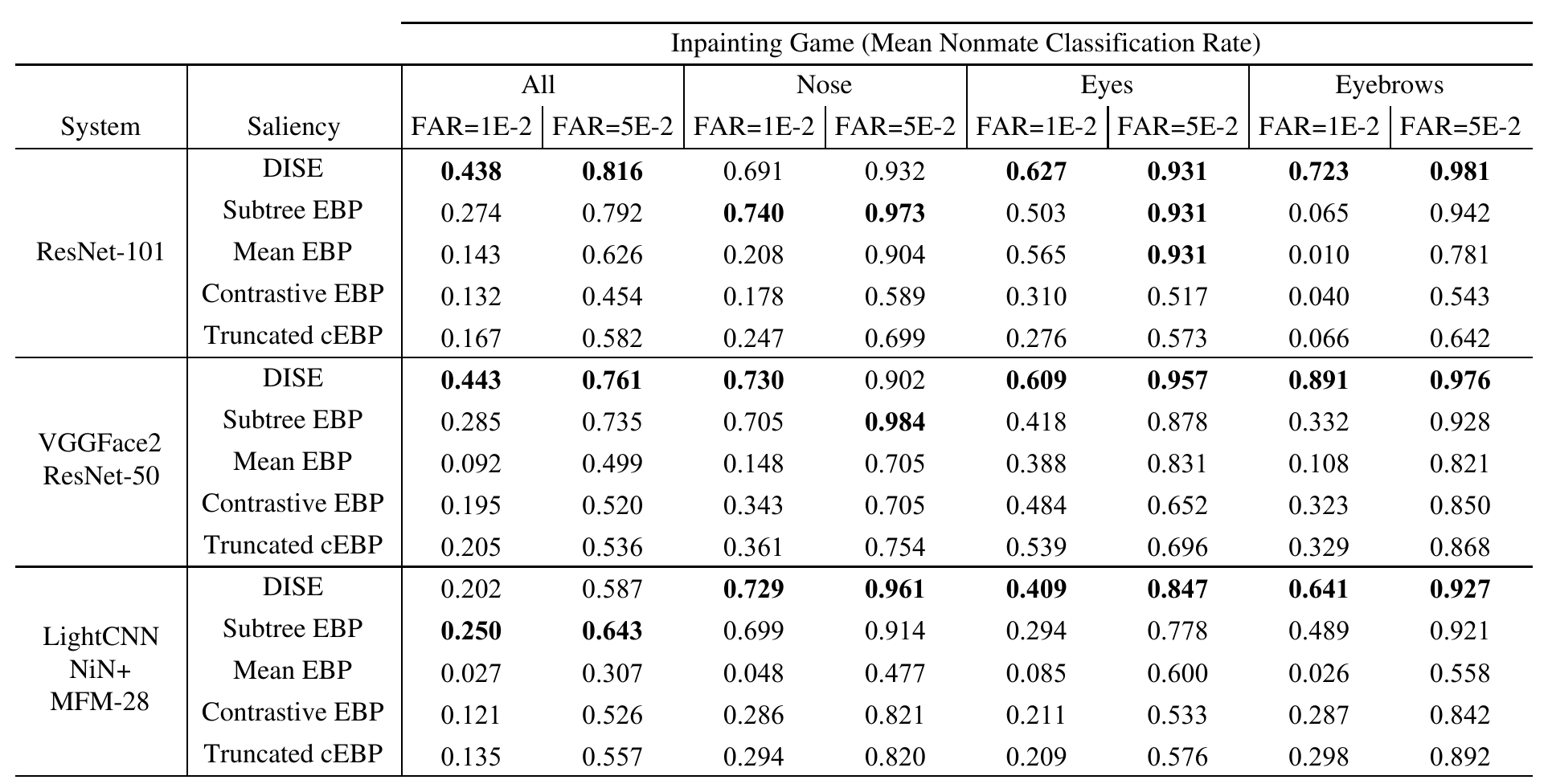}
    \caption{
    Inpainting game evaluation results.
    This table summarizes the performance at two operating points of false alarm rate (1E-2, 5E-2) for the performance curves in Fig. \ref{f:inpainting_game_results_plots} (ResNet-50 and LightCNN) and 
    in the supplementary material. 
    Mean nonmate classification rate is the proportion of triplets where the identity of the blended image was correctly ``flipped'' to the doppelganger. 
    Results show that our new methods (DISE, Subtree EBP) outperform the state of the art by a wide margin on three matchers. Detailed subprotocol results and curves are provided in the supplemental material.
    }
     \label{t:inpainting_game_results_summary}
\end{table}

\section{Experimental Results}
\label{s:results}

\subsection{Inpainting Game Quantitative Evaluation}
\label{s:inpainting_game_results}

We ran the inpainting game evaluation protocol on the inpainting dataset using three target networks:  LightCNN \cite{wu2018light}, VGGFace2 ResNet-50 \cite{Cao18} and a custom trained ResNet-101.  We considered the five XFR algorithms described in Sec. \ref{s:xfr} forming the benchmark for XFR evaluation.

The evaluation results are summarized in Table \ref{t:inpainting_game_results_summary} and plotted in Fig. \ref{f:inpainting_game_results_plots} and 
in the supplementary material.
The summary table shows for each combination of network and XFR algorithm, at two false alarm rates (1E-2, 5E-2) for the full protocol and three subprotocols: eyes, nose and eyebrows only.  
Additional results in the supplementary material 
show the results for the individual facial region subprotocols.  

Overall, results show that for deeper networks (ResNet-101, ResNet-50), the top performing XFR algorithm is DISE.  However, for shallower networks (LightCNN) then top performing algorithm is Subtree EBP.  Both of these new approaches outperform the state of the art (EBP, cEBP, tcEBP) by a wide margin.  We believe that DISE outperforms Subtree EBP since subtree EBP cannot localize image regions any better than the underlying network represents faces.  For example, consider the eyebrows subprotocol result in 
the supplementary material,
which shows that subtree EBP cannot represent eyebrows independently from the eyes.  DISE can mask image regions independently from the underlying target network and correctly localize eyebrow effects.

\section{Conclusions}

In this paper, we introduced the first comprehensive benchmark for XFR  We motivated the need for XFR and describe a new quantitative method for comparing XFR algorithms using the inpainting game.  The results show that the DISE and subtree EBP methods provide a significant performance improvement over the state of the art, which provides a new baseline for visualizing discriminative features for face recognition.  This evaluation protocol provides a means to compare different approaches to network saliency, and we believe this form of quantitative evaluation will help encourage research in this emerging area of explainable AI for face recognition.  All software and datasets for reproducible research are available for download at \url{http://stresearch.github.io/xfr}.

\medskip
\noindent {\bf Acknowledgement.}
This research is based upon work supported by the Office of the Director of National Intelligence (ODNI), Intelligence Advanced Research Projects Activity (IARPA) under contract number 2019-19022600003. The views and conclusions contained herein are those of the authors and should not be interpreted as necessarily representing the official policies or endorsements, either expressed or implied, of ODNI, IARPA, or the U.S. Government.  The U.S. Government is authorized to reproduce and distribute reprints for Governmental purpose notwithstanding any copyright annotation thereon.

\bibliographystyle{ieee}
\bibliography{bib/jebyrne,bib/Subjectness,bib/janus}

\appendix
\onecolumn
\section{Supplementary Material}
\beginsupplement
\subsection{Qualitative Visualization Study}
\label{s:qualitative}

\begin{figure*}[t]
    \centering
    \includegraphics[width=\linewidth]{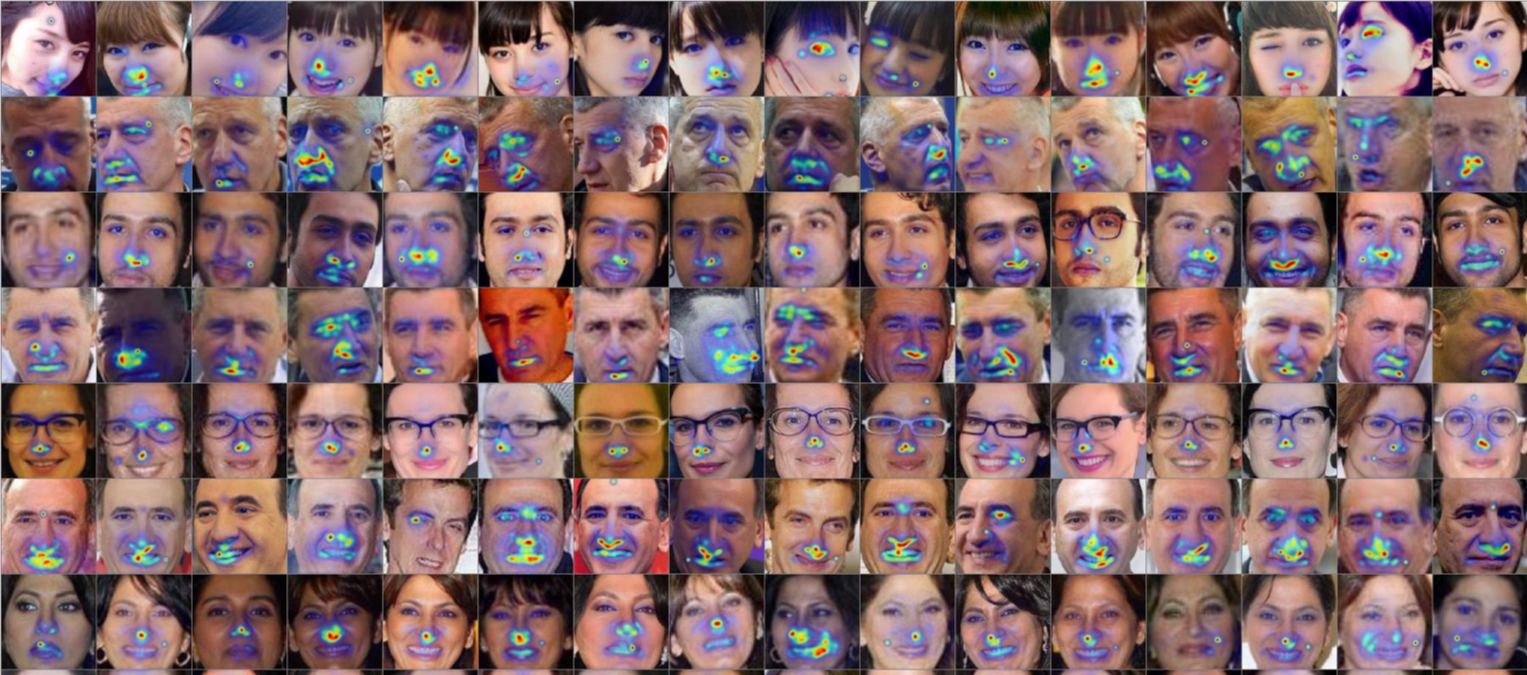}
    \caption{Qualitative visualization study.  This figure shows the XFR saliency maps generated using the LightCNN Subtree EBP method for 16 probes (columns) of 7 subjects (rows), each with 16 mates (not shown) and a common set of 8000 nonmates (not shown) all sampled from VGGFace2 \cite{Cao18}.  Results show that the discriminative features used to distinguish a subject from the entire nonmate population are inconsistent, but are primarily the nose and mouth for frontal probes, including the eyes for non-frontal probes. 
    See supplemental Fig. \ref{f:whitebox_onevsrest} for additional examples. 
    }
     \label{f:whitebox_onevsrest_sevenrows}
\end{figure*}

\begin{figure*}
    \centering
    \includegraphics[width=\linewidth]{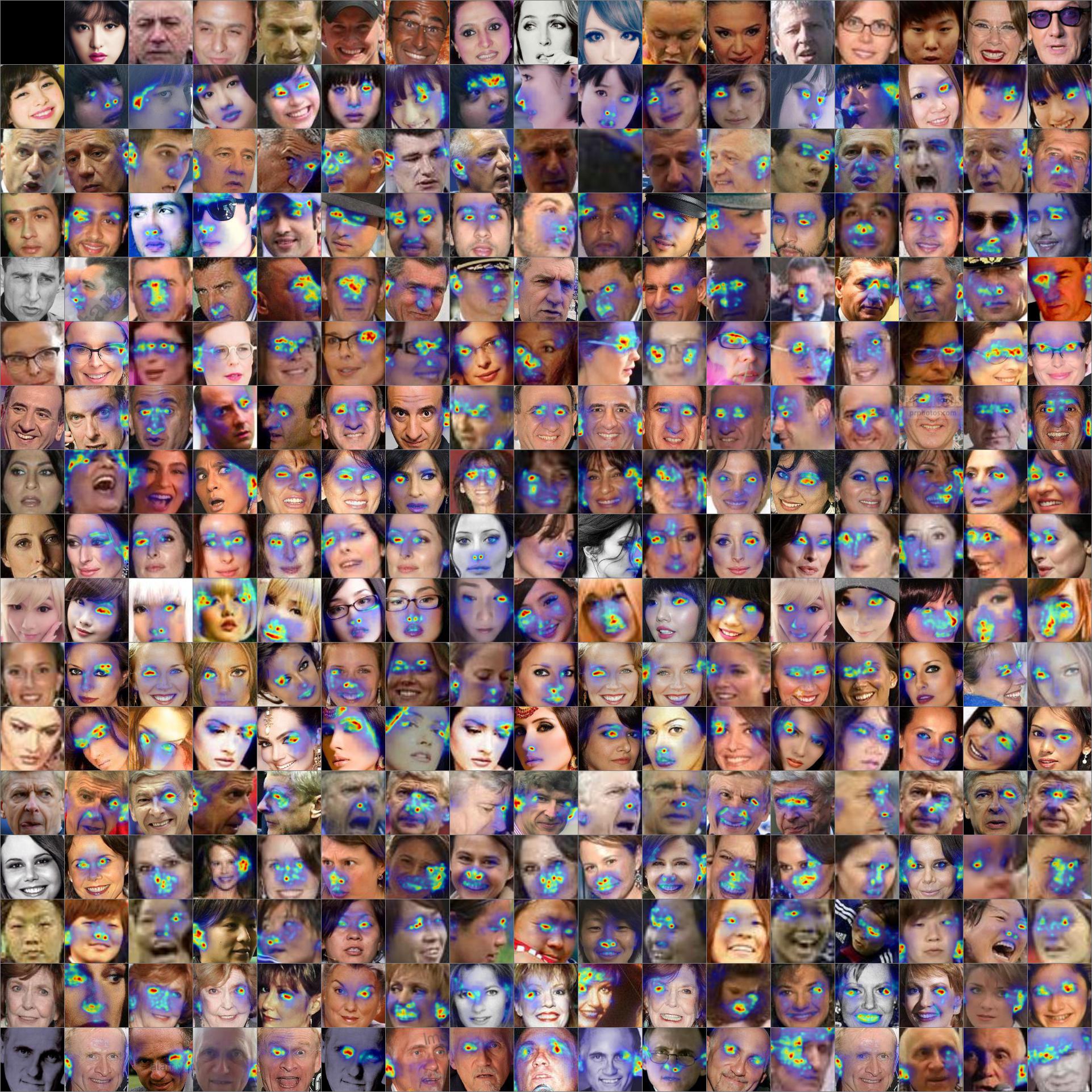}
    \caption{Contrastive triplet EBP (ResNet-101).  This montage shows the contrastive EBP 
    for a ResNet-101 whitebox.  Observe that this saliency map is unstable, and at times generates saliency maps on the background of the image (e.g. probe (16,15)).  This is a known challenge of contrastive EBP, which led towards the development of truncated contrastive EBP.
    }
     \label{f:cebp_str}
\end{figure*}

\begin{figure*}
    \centering
    \includegraphics[width=\linewidth]{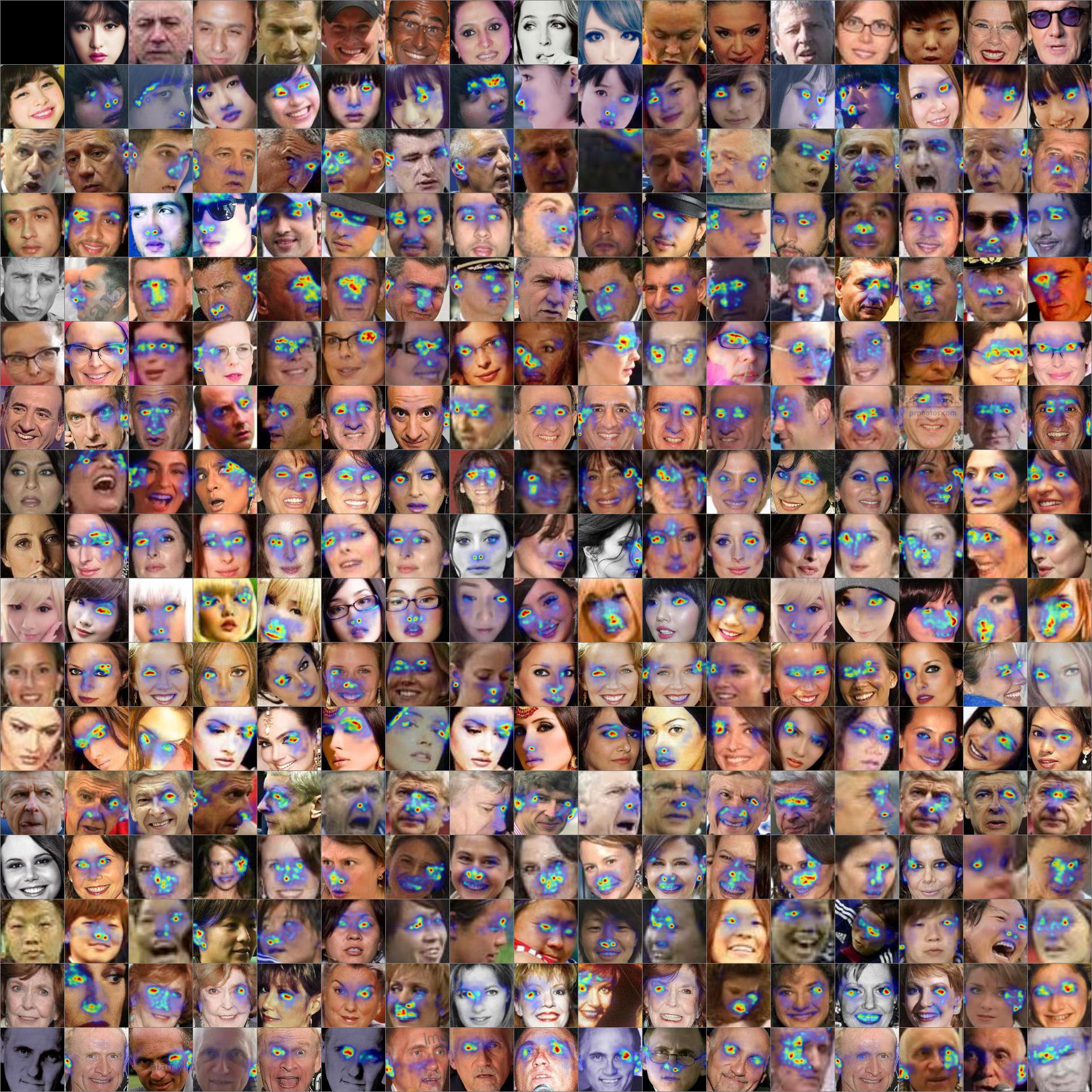}
    \caption{Truncated contrastive triplet EBP (ResNet-101).  This montage shows truncated contrastive triplet EBP.
    }
     \label{f:tcebp_str}
\end{figure*}

\begin{figure*}
    \centering
    \includegraphics[width=\linewidth]{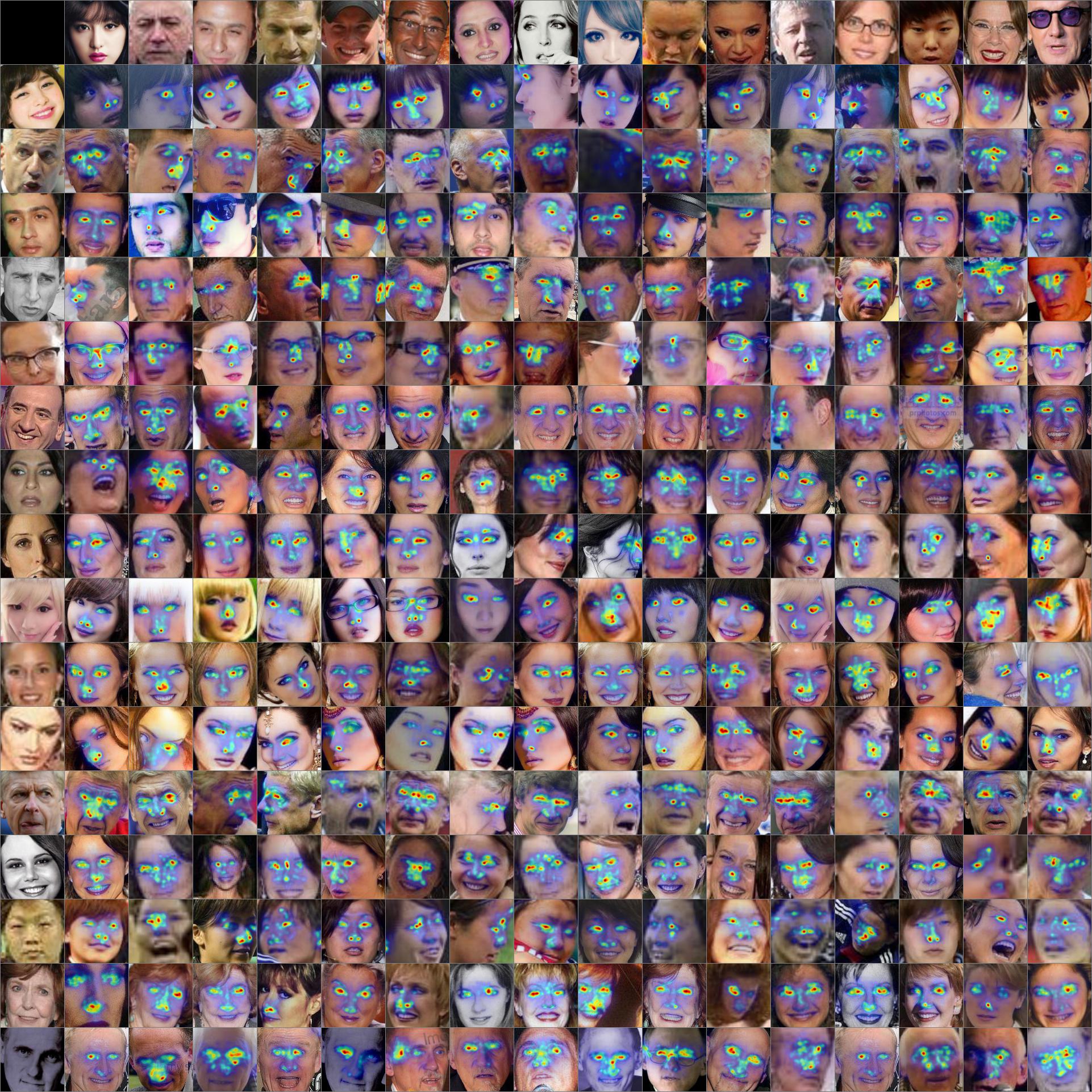}
    \caption{Subtree Triplet EBP (ResNet-101).  This montage shows subtree triplet EBP. 
    }
     \label{f:subtree_str}
\end{figure*}

\begin{figure*}
    \centering
    \includegraphics[width=\linewidth]{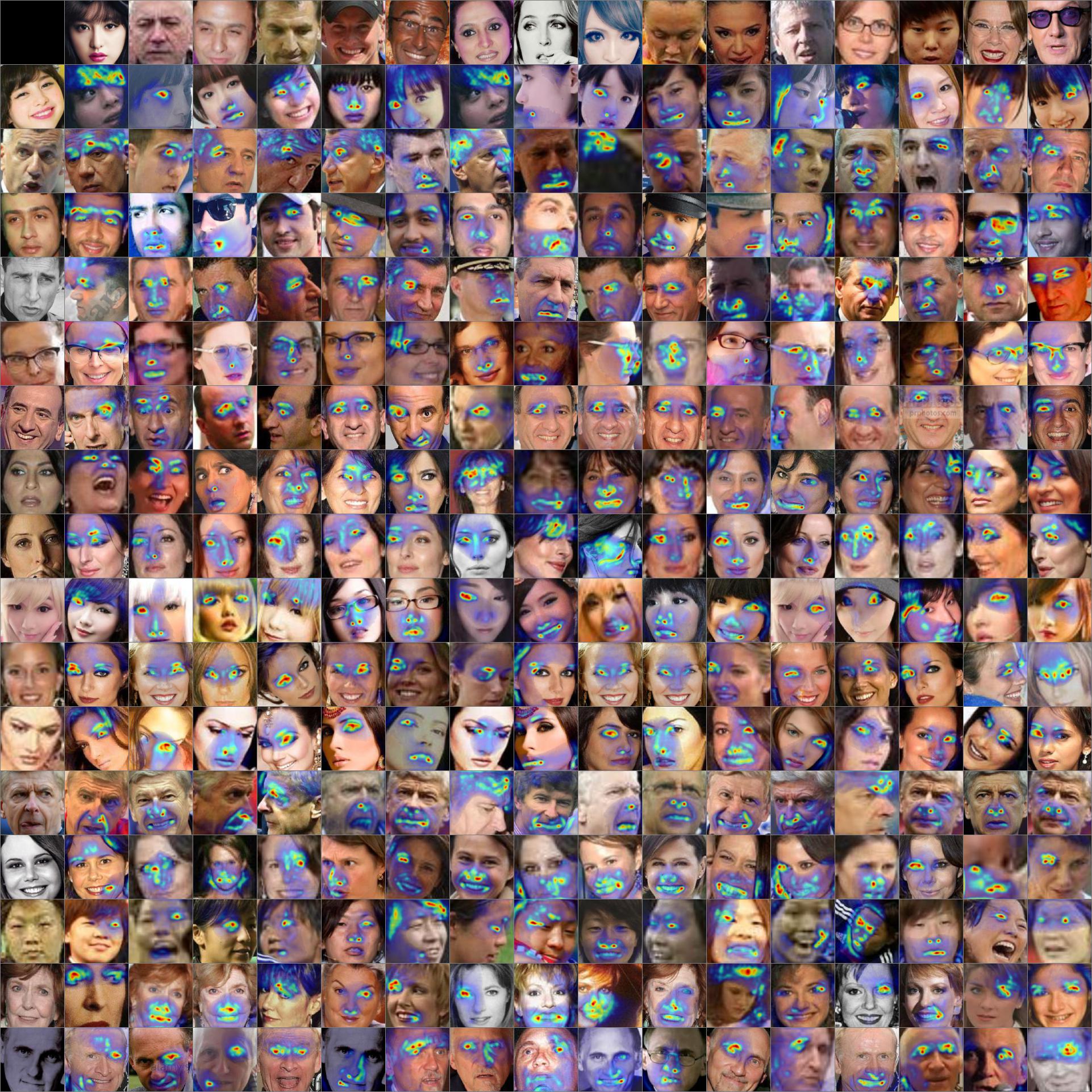}
    \caption{ Contrastive triplet EBP (Light-CNN \cite{wu2018light}).  This montage should be compared with Fig. \ref{f:cebp_str} to show the differences for contrastive triplet EBP comparing ResNet-101 with light-CNN.
    }
     \label{f:cebp_lightcnn}
\end{figure*}

\begin{figure*}
    \centering
    \includegraphics[width=\linewidth]{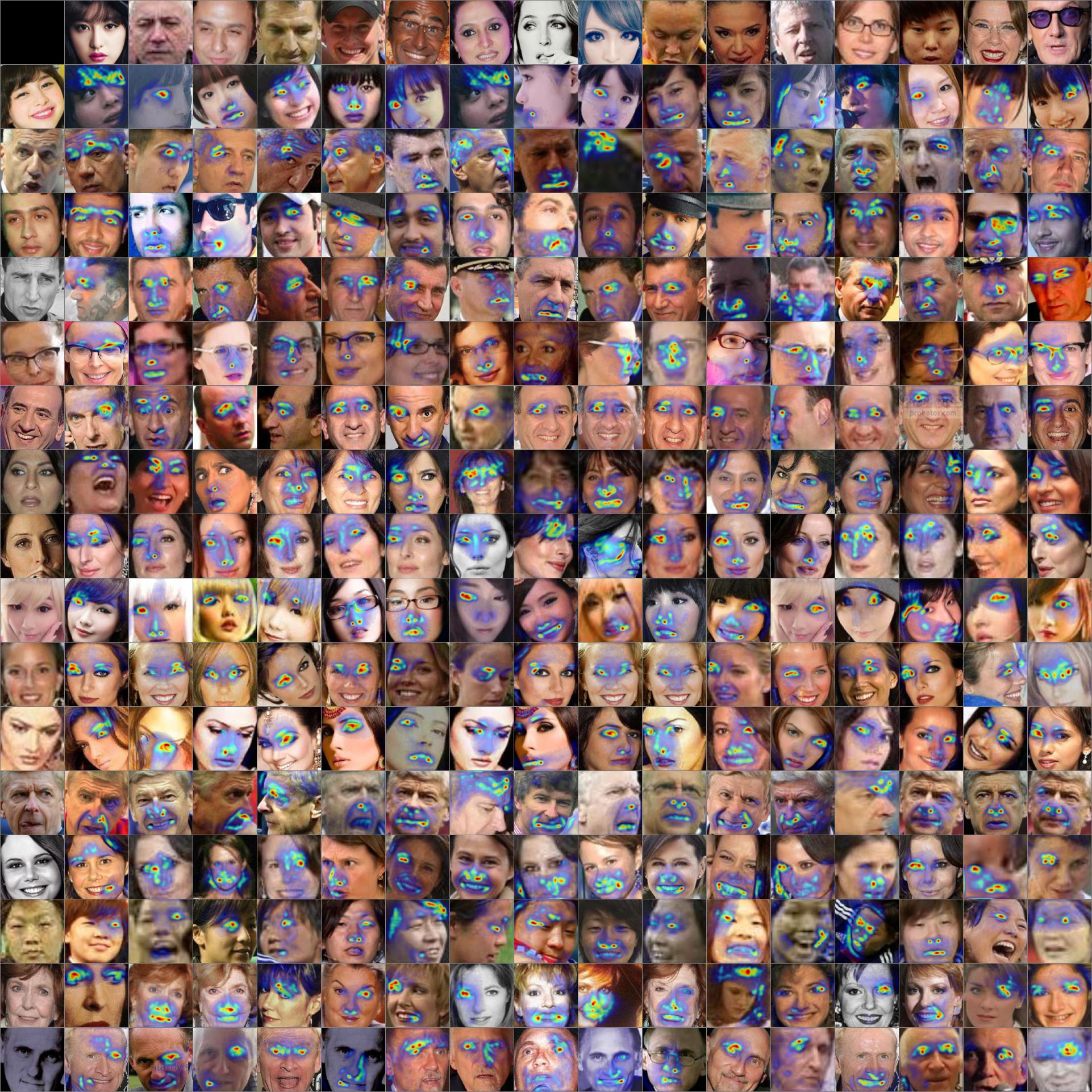}
    \caption{Truncated contrastive triplet EBP (Light-CNN \cite{wu2018light}).  This montage should be compared with Fig. \ref{f:tcebp_str} to show the differences for tcEBP comparing ResNet-101 with light-CNN.
    }
     \label{f:tcebp_lightcnn}
\end{figure*}

\begin{figure*}
    \centering
    \includegraphics[width=\linewidth]{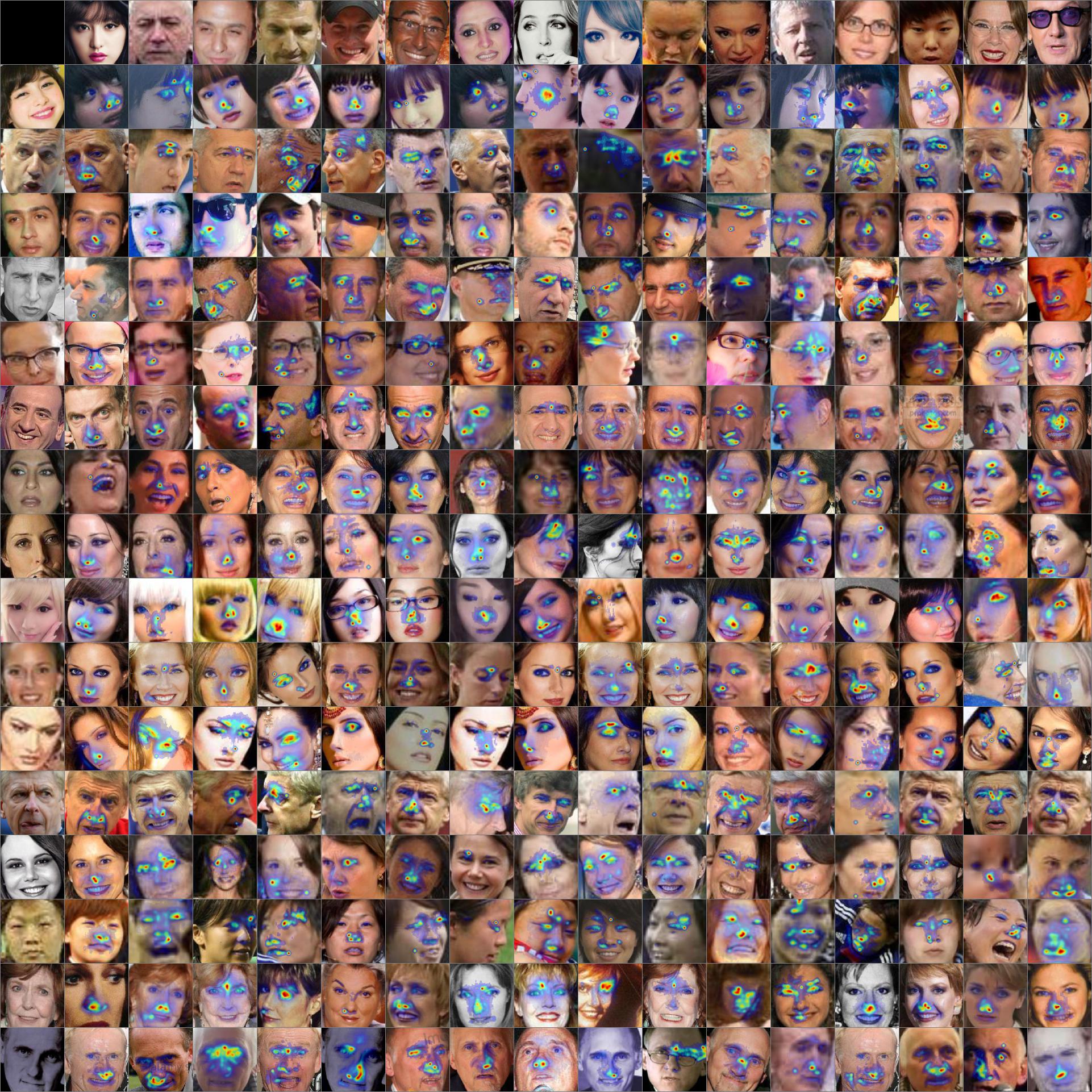}
    \caption{Subtree triplet EBP (Light-CNN \cite{wu2018light}).  This montage should be compared with Fig. \ref{f:subtree_str} to compare the differences for subtree triplet EBP for ResNet-101 vs. light-CNN.
    }
     \label{f:subtree_lightcnn}
\end{figure*}

\begin{figure*}
    \centering
    \includegraphics[width=\linewidth]{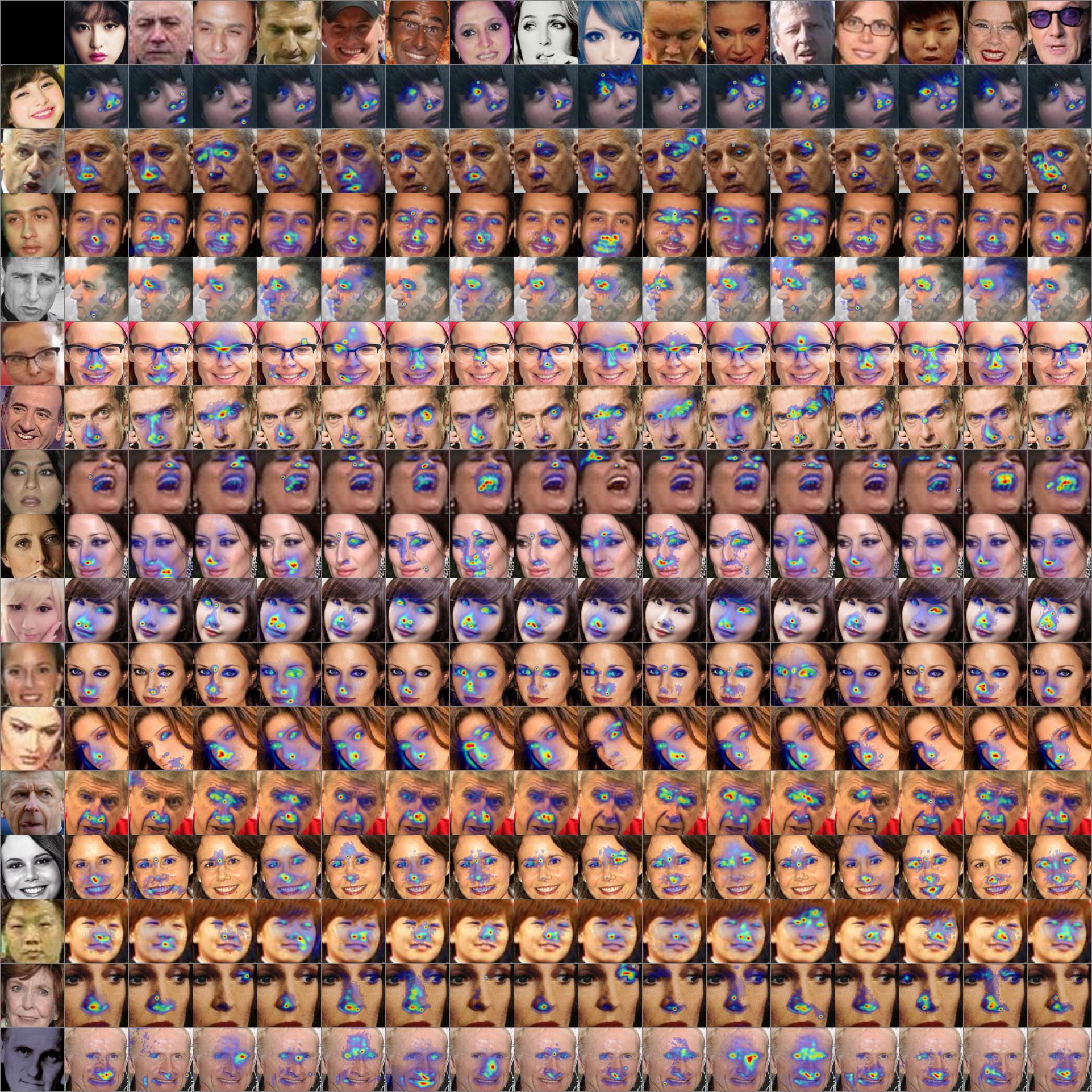}
    \caption{Repeated probe montage (Light-CNN \cite{wu2018light}).  This montage shows the same probe repeated across each row to highlight the effect of the non-mate in the triplet on the resulting saliency map.    
    }
     \label{f:repeated_probe_montage_lightcnn}
\end{figure*}

The inpainting game provides a quantitative comparison of XFR algorithms, however it does not provide insight as to how useful these XFR algorithms are on novel face images.  In this section, we provide a qualitative study of XFR algorithms visualized on a standard set of triplets. 
We consider two target networks: ResNet-101 and Light-CNN \cite{wu2018light}, and provide visualizations for the whitebox methods referenced in 
the main submission.
This analysis includes the following figures showing qualitative visualization results for combinations of (target network, XFR method): (ResNet-101, EBP, Fig. \ref{f:ebp_str}), (ResNet-101, cEBP,  Fig. \ref{f:cebp_str}), (ResNet-101, tcEBP, Fig. \ref{f:tcebp_str}), (ResNet-101, Subtree, Fig. \ref{f:subtree_str}), (Light-CNN, EBP, Fig. \ref{f:ebp_lightcnn}), (Light-CNN, cEBP, Fig. \ref{f:cebp_lightcnn}), (Light-CNN, tcEBP, Fig. \ref{f:tcebp_lightcnn}), (Light-CNN, Subtree EBP, Fig. \ref{f:subtree_lightcnn}).  Finally, we show results for the Light-CNN using only single probes (Fig. \ref{f:single_probe_montage_lightcnn}), or repeated probes (Fig. (\ref{f:repeated_probe_montage_lightcnn}) to highlight the effect of non-mates in the triplet visualization.

From this visualization study, we draw the following conclusions:

\begin{enumerate}
    \item {\bf Non-localized.}  
    Unlike facial examiners which leverage the complete FISWG standards for facial comparison, there is no evidence that modern face matchers leverage localized discriminating features such as scars, marks and blemishes.  All visualizations are centered on the facial interior, and almost no activation is on the shape of the head.  Also, the systems tend to overgeneralize to represent all faces in a standard manner using the eyes and nose, brow and mouth, ignoring localized features such as moles or facial markings.    
    \item {\bf Pose variant.}  The target networks tested are not truly pose invariant.  When considering different probes of the same subject, where the probe differs in pose, the whitebox systems can generate different visualizations.  This suggests that the underlying network is still pose variant.
    \item {\bf Triplet specific.}  The features that are used for recognition depend on the selection of the triplet, notably the selection of the non-mate for comparison.  The visualized features are more consistent when considering a larger non-mate set (Fig. \ref{f:whitebox_onevsrest_sevenrows}).
    \item {\bf Network specific.}  The visualized features are dependent on the selected target network for visualization.  A higher performing network (light-CNN) tends to use more facial features of the brow and mouth in addition to the eyes and nose, than a lower performing network (ResNet-50). No networks yet tested use the hair or chin.  
\end{enumerate}

\begin{figure*}
    \centering
    \includegraphics[width=\linewidth]{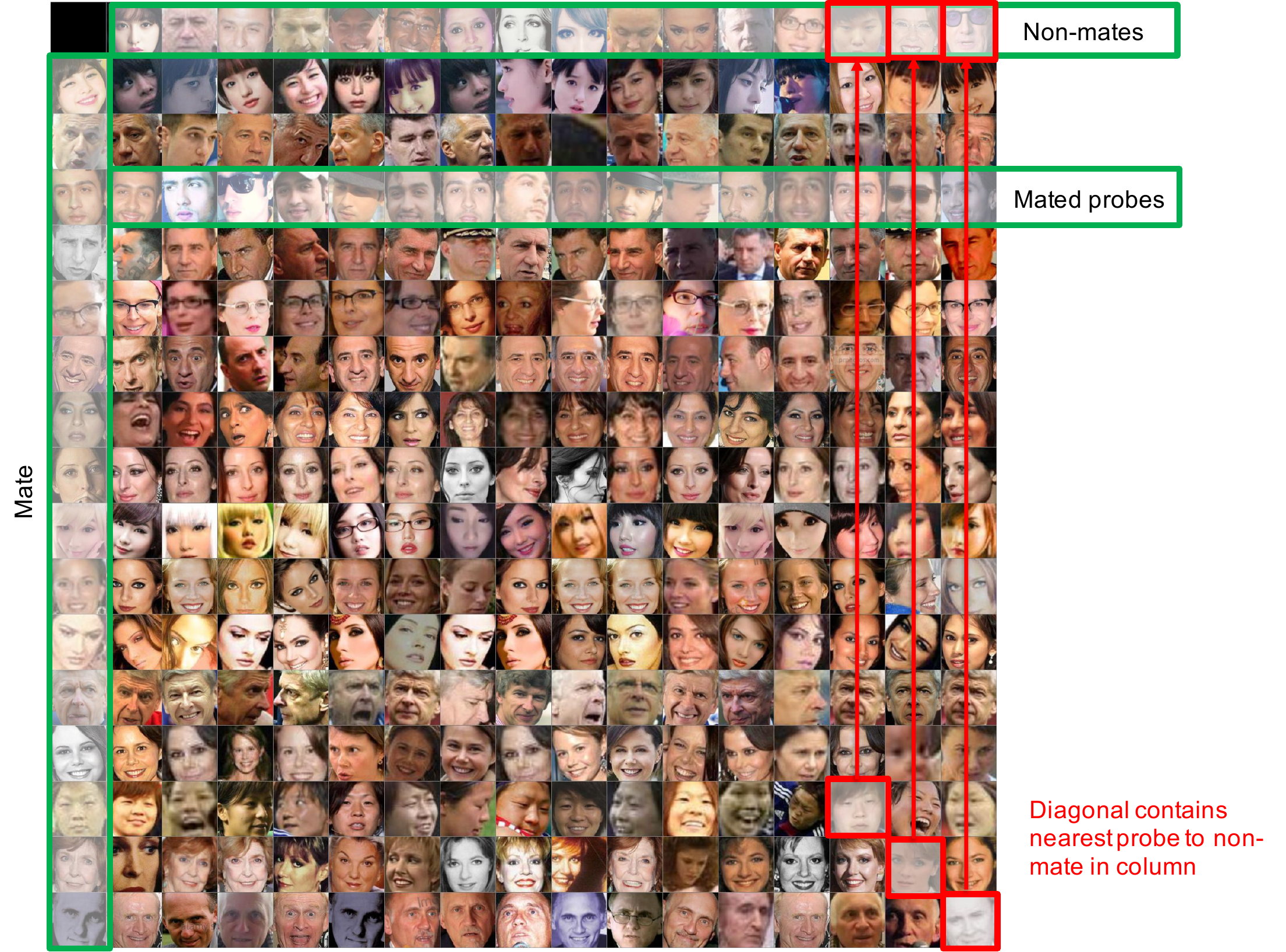}
    \caption{Whitebox visualization overview.  This montage shows a set of 16 randomly selected subjects from IJB-C, such that every row has the same identity.  Images $(i,j)$ in this montage define a triplet $(m_i,p_{ij},n_j)$ for probe $p_{ij}$, mate $m_i$ in the first entry of column $i$ and non-mate $n_j$ in the first entry in row $j$.  Non-mates are ordered such that on the diagonal are the nearest non-mated subject in IJB-C. In other words, for triplet $(m_i,p_{ii},n_i)$, non-mate $n_i$ is more similar to $m_i$ than any other nonmate $n_j$, using a ResNet-101 matching system.  This montage is used to visualize how the whitebox saliency map changes when considering different triplets.   
    }
     \label{f:whitebox_visualization_overview}
\end{figure*}

\begin{figure*}
    \centering
    \includegraphics[width=\linewidth]{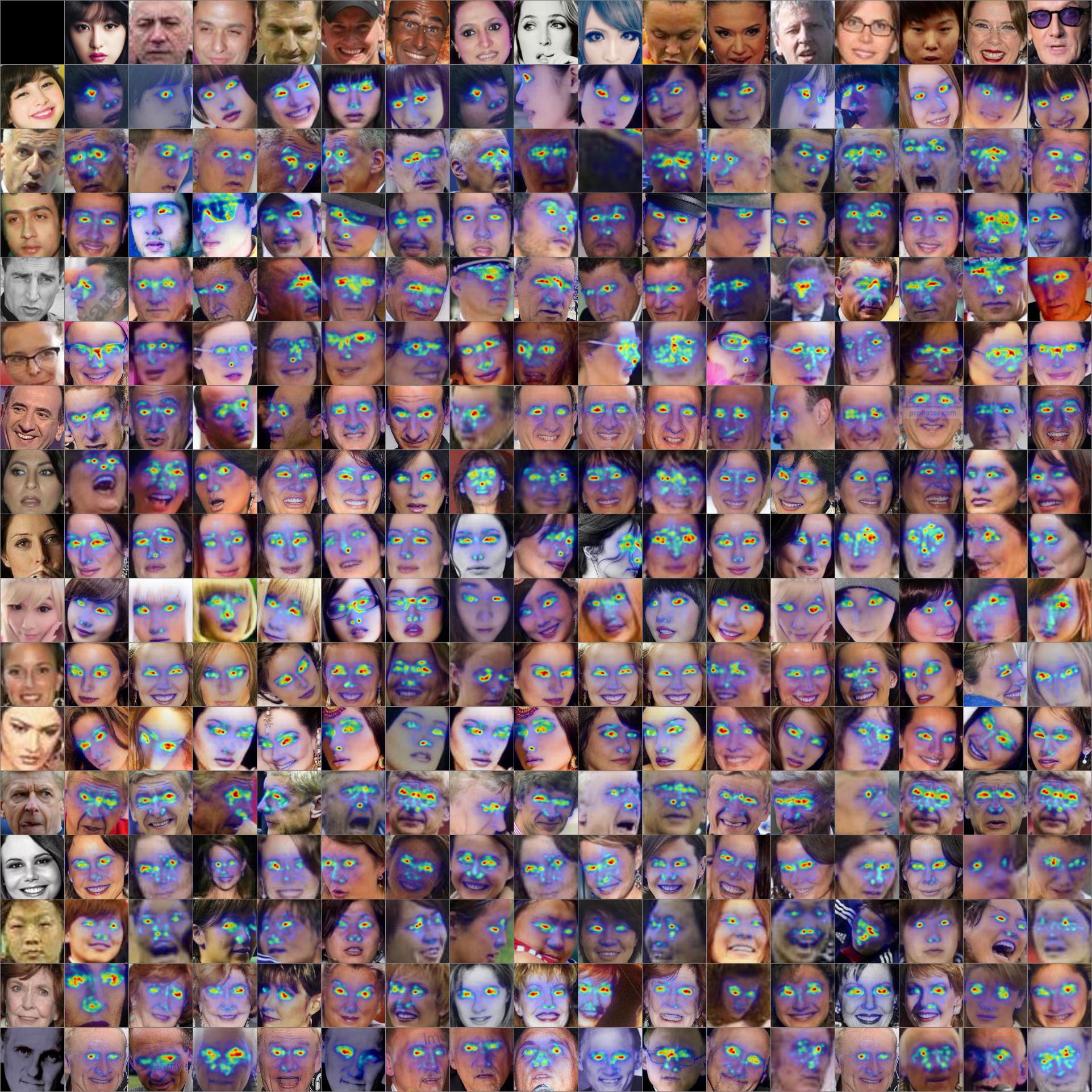}
    \caption{EBP (ResNet-101).  This montage is the same images as in Fig. \ref{f:whitebox_visualization_overview}, but with a whitebox saliency map derived from excitation backprop for a whitebox ResNet-101 system. 
    Observe that EBP always selects the eyes and nose no matter what non-mated subject is being considered.  This does not provide subtle distinctions between the regions that are discriminative for a mate vs. a non-mate, but it does provide a visualization of the regions of the probe that are used for classification.  This visualization should be compared with Fig. \ref{f:ebp_lightcnn} for the same subjects and whitebox method, but a different underlying trained network (light-cnn).
    }
     \label{f:ebp_str}
\end{figure*}

\begin{figure*}
    \centering
    \includegraphics[width=\linewidth]{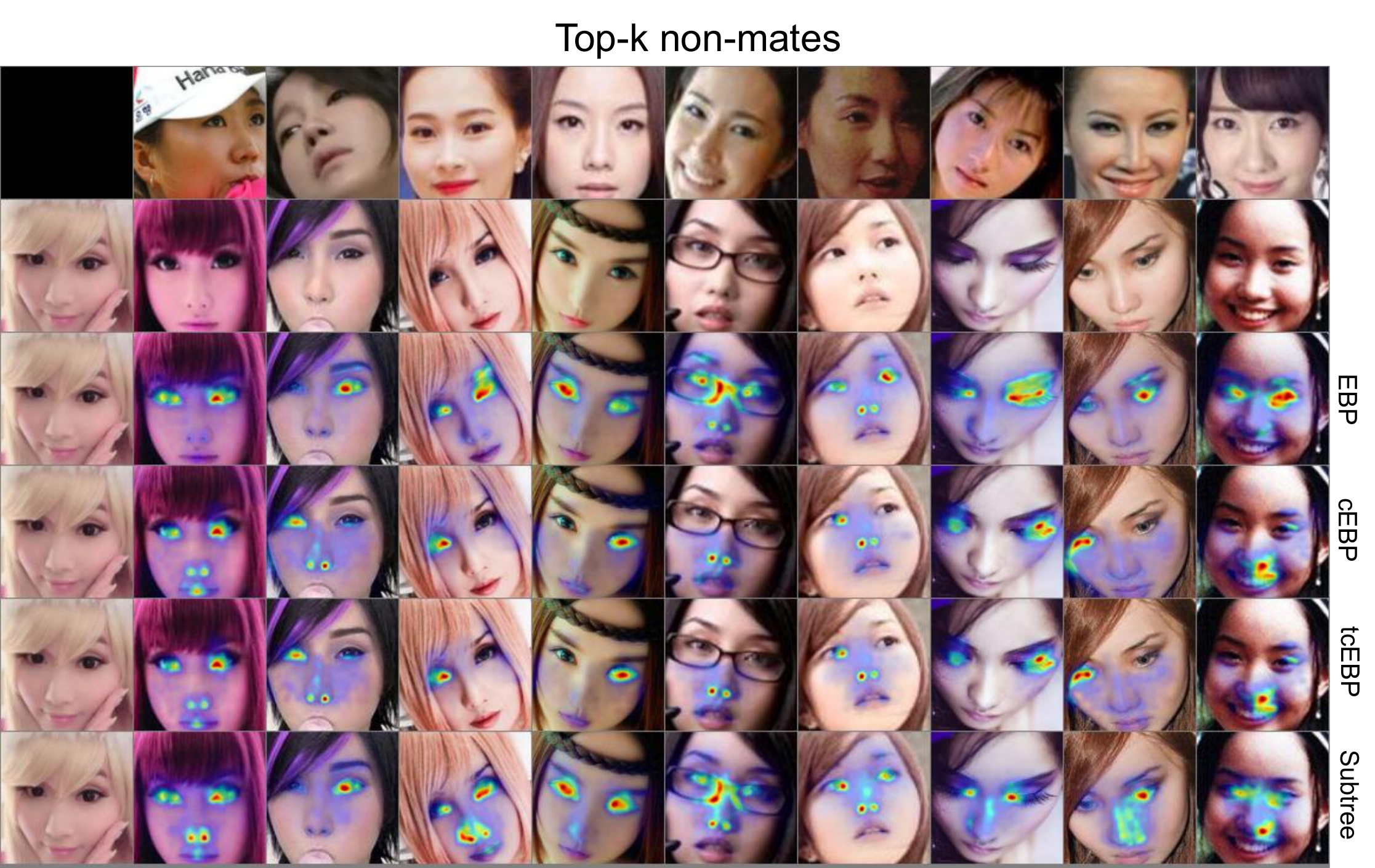}
    \caption{Single probe montage (ResNet-101).  This montage compares four white box methods on a common set of probes, such that the non-mates are now ordered in decreasing similarity with the mate.  This shows how the different methods compare for real-world doppelgangers.  
    }
     \label{f:montage_str}
\end{figure*}

\begin{figure*}
    \centering
    \includegraphics[width=\linewidth]{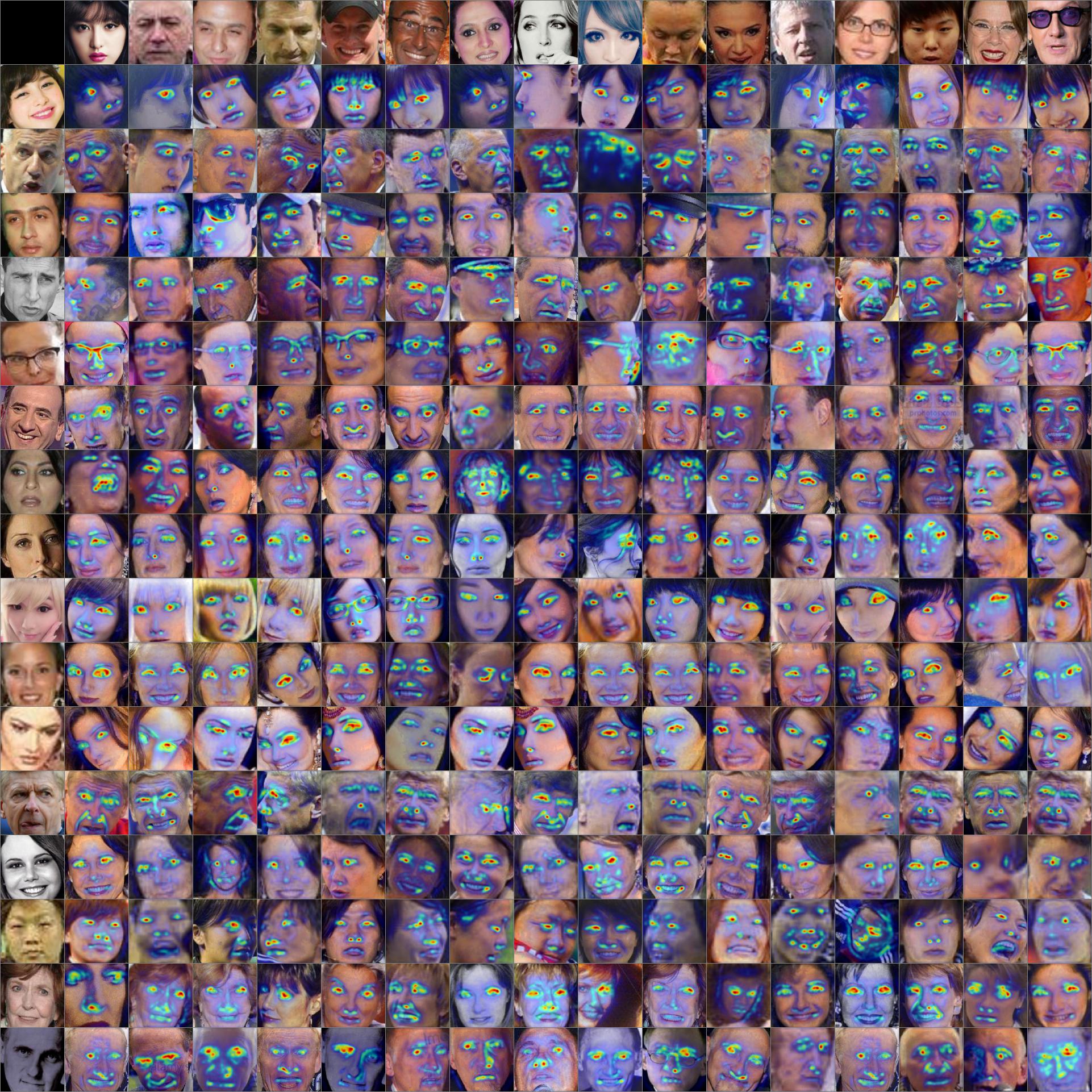}
    \caption{EBP (Light-CNN \cite{wu2018light}).  This montage generates the EBP saliency map for the Light-CNN network.  This should be compared with Fig. \ref{f:ebp_str}, which shows that this network exhibits more saliency around the mouth and brow than the ResNet-101 network.  
    }
     \label{f:ebp_lightcnn}
\end{figure*}

\begin{figure*}
    \centering
    \includegraphics[width=\linewidth]{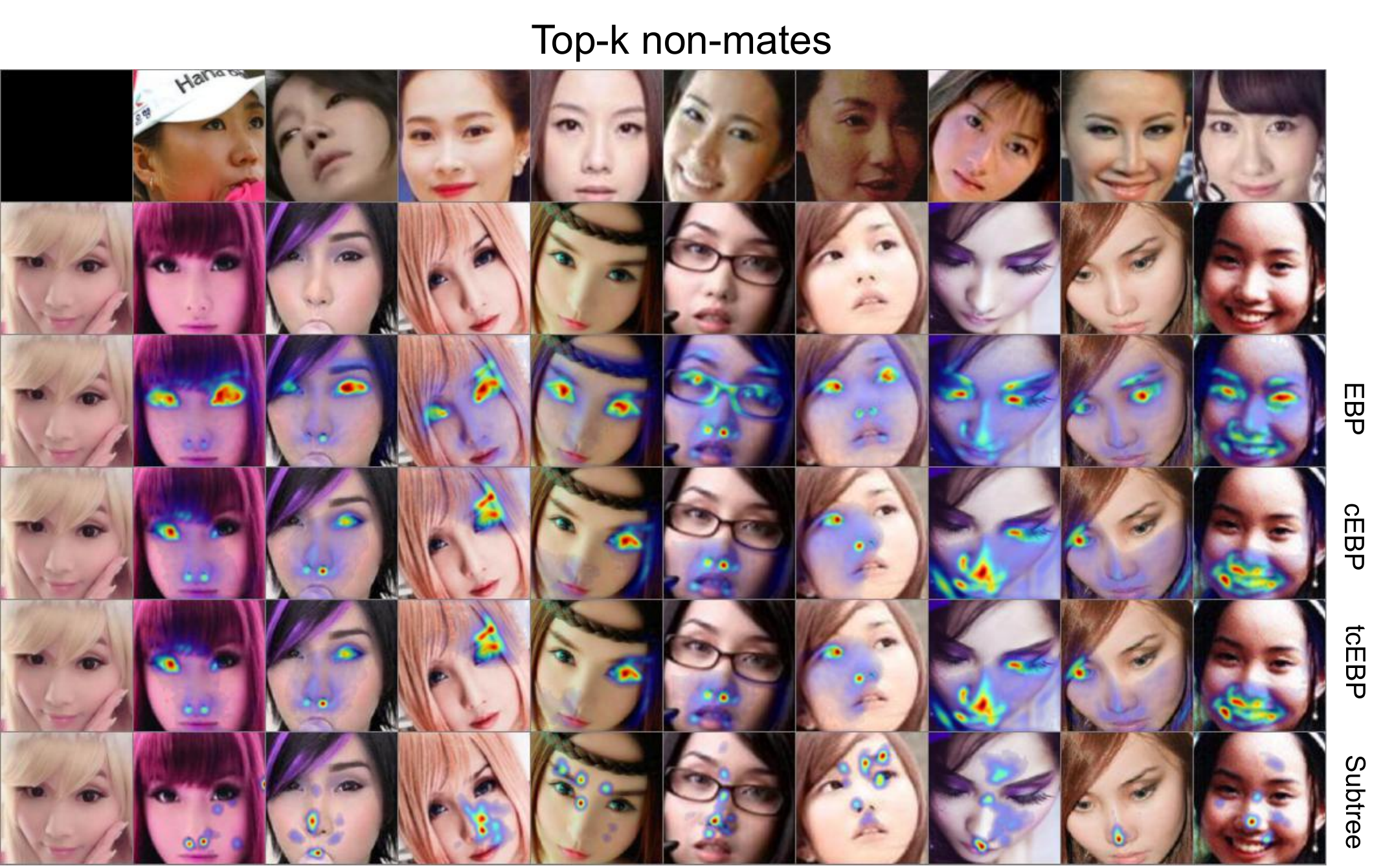}
    \caption{Single probe montage (Light-CNN \cite{wu2018light}).  This montage should be compared with Fig. \ref{f:montage_str} to compare the effect of top-k non-mates for ResNet-101 vs. light-CNN.
    }
     \label{f:single_probe_montage_lightcnn}
\end{figure*}

\begin{figure*}
    \centering
    \includegraphics[width=\linewidth]{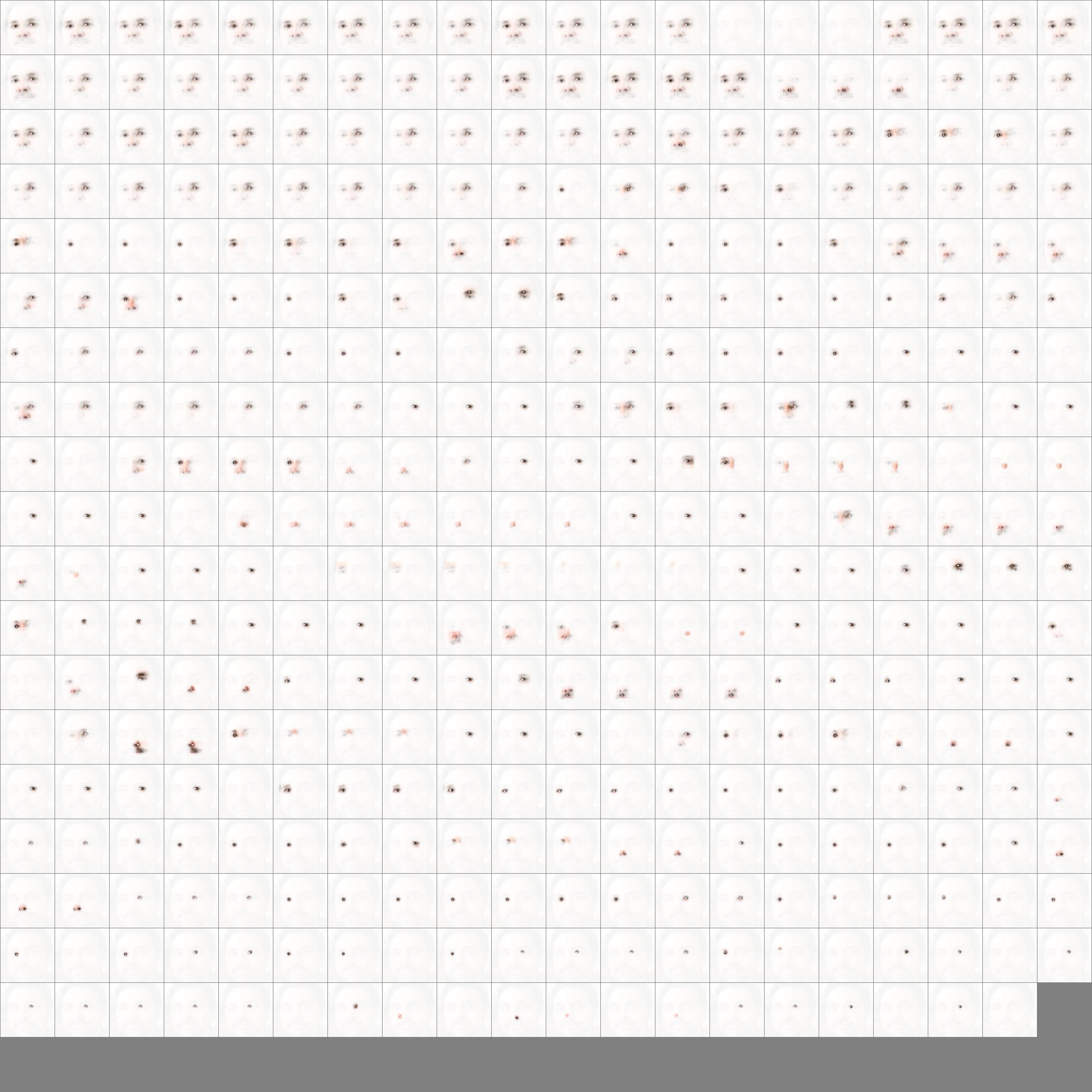}
    \caption{Layerwise EBP.  This montage shows the EBP saliency map generated starting from the maximum excitation for each layer in a ResNet-101 network.  The layers are ordered rowwise, starting from the embedding layer in the upper left down to the image layer in the bottom right.  The visualization shows the saliency map encoded as the alpha channel of a cropped face image, so that non-zero saliency results in a more opaque (less transparent) region.  This visualization style is useful to accentuate small activations.  This result shows that saliency maps starting from the layers closer to the embedding result in holistic regions covering the eyes and nose, layers in the middle show parts such as the eyes, nose and mouth, layers closer to the image are highly localized on specific regions of the image, and some layers provide no excitation at all.    
    }
     \label{f:layerwise_ebp}
\end{figure*}

\begin{figure*}
    \centering
    \includegraphics[width=\linewidth]{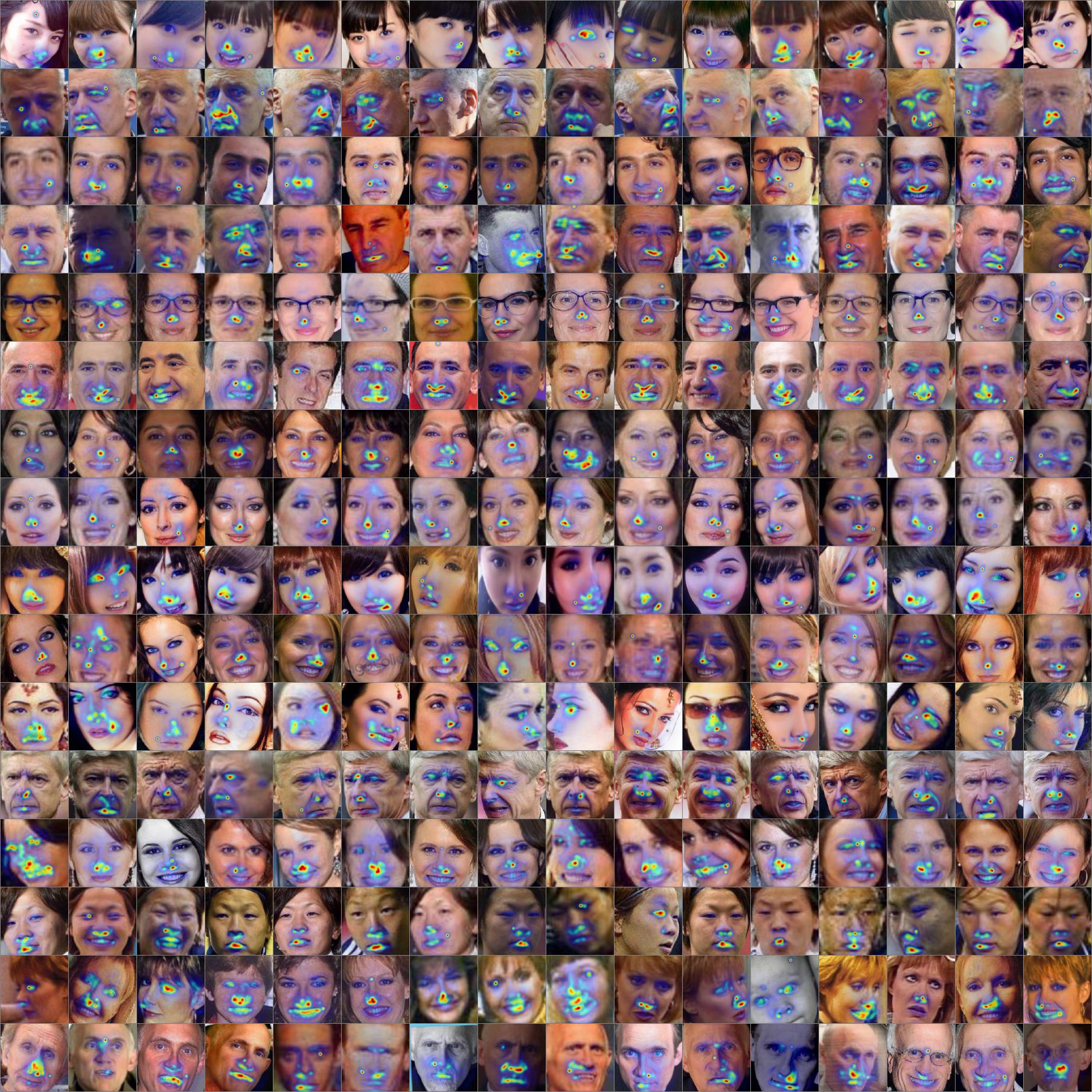}
    \caption{Qualititative visualization study.  This figure shows the XFR saliency maps generated using the LightCNN Subtree EBP method for 16 probes (columns) of 16 subjects (rows), each with 16 mates (not shown) and a common set of 8000 nonmates (not shown) all sampled from VGGFace2 \cite{Cao18}.  Results show that the discriminative features used to distinguish a subject from the entire nonmate population are primarily the nose and mouth for frontal probes and eyes for non-frontal probes.  These network attention maps are remarkably consistent across probes and provide insight into the features that a network uses to distinguish a subject from a large set of nonmates (i.e. What makes you unique?).}
     \label{f:whitebox_onevsrest}
\end{figure*}


\begin{figure*}
    \centering
    \includegraphics[width=\linewidth]{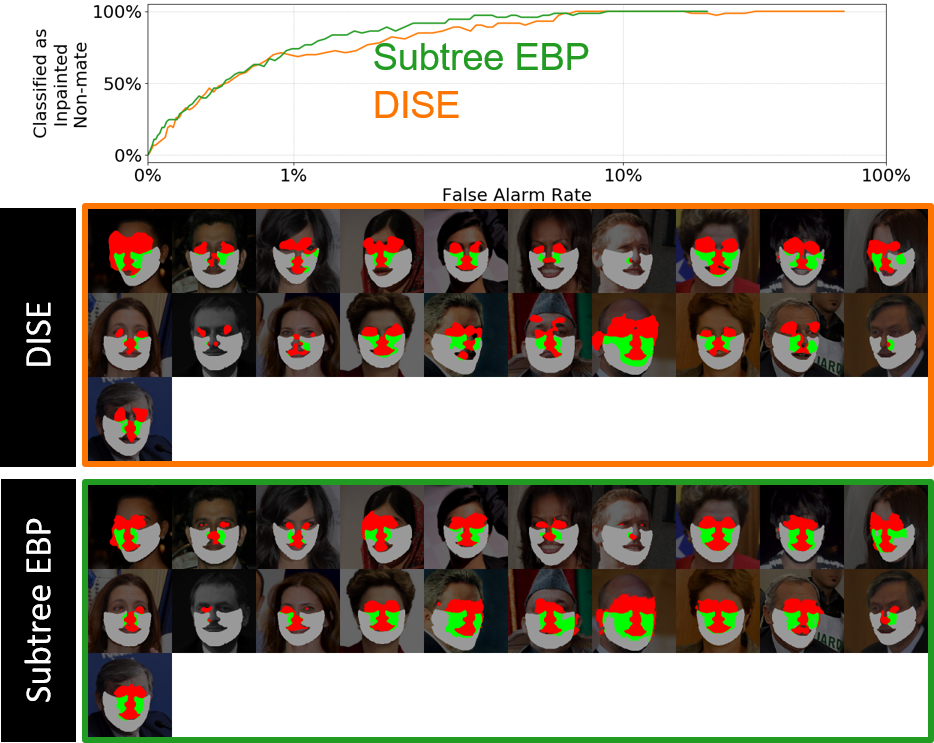}
    \caption{
Cheek/Chin Mask (ResNet-101): Evaluation plot and classification on saliency maps from
Subtree EBP  and
DISE  at identity flip.
}
   \label{fig:ResNetv4_mask0} 
\end{figure*}


\begin{figure*}
    \centering
    
    \includegraphics[width=\linewidth]{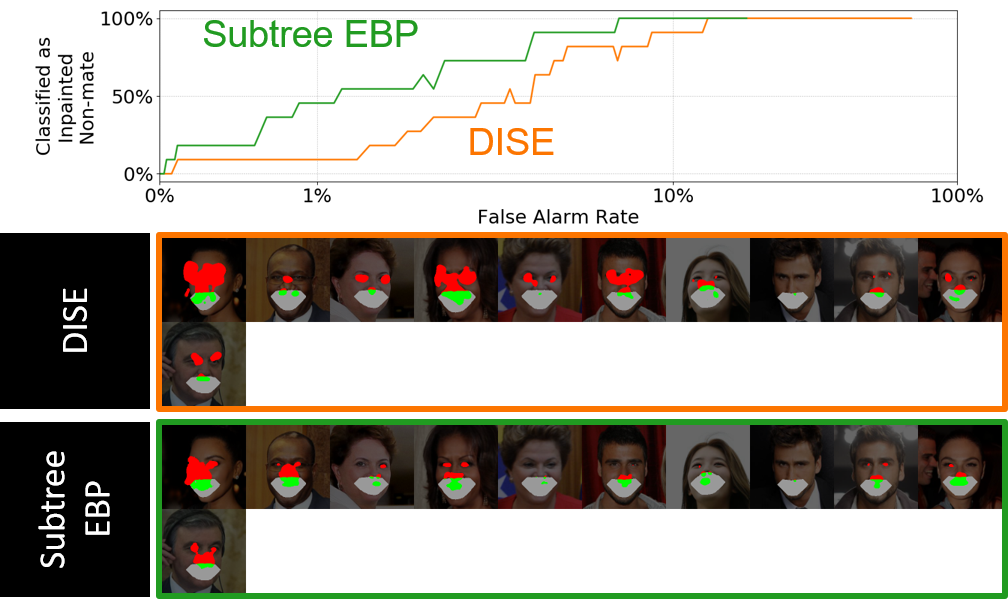}
    \caption{Mouth Mask (ResNet-101): Evaluation plot and classification on saliency maps from
    Subtree EBP  and
    DISE  at identity flip.}
\label{fig:ResNetv4_mask1} 
\end{figure*}


\begin{figure*}
    \centering
    \includegraphics[width=\linewidth]{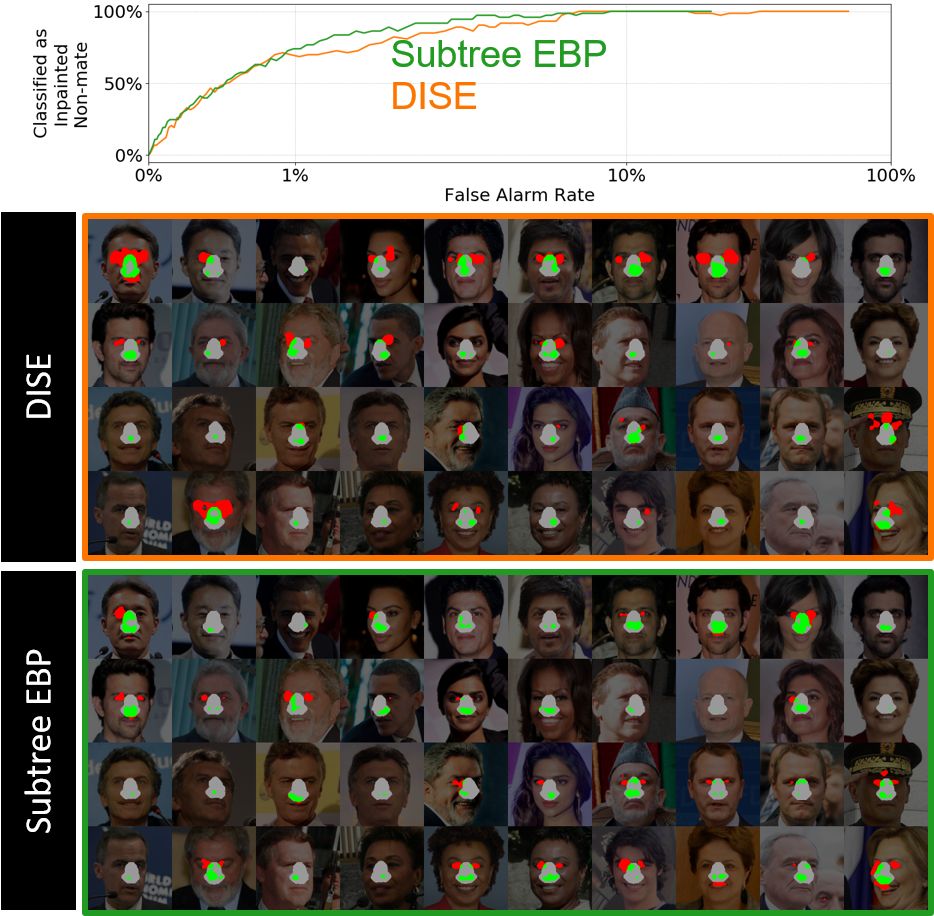}
    \caption{Nose Mask (ResNet-101): Evaluation plot and classification on saliency maps from
    Subtree EBP  and
    DISE  at identity flip.}
\label{fig:ResNetv4_mask2} 
\end{figure*}

\begin{figure*}
    \centering
    \includegraphics[width=\linewidth]{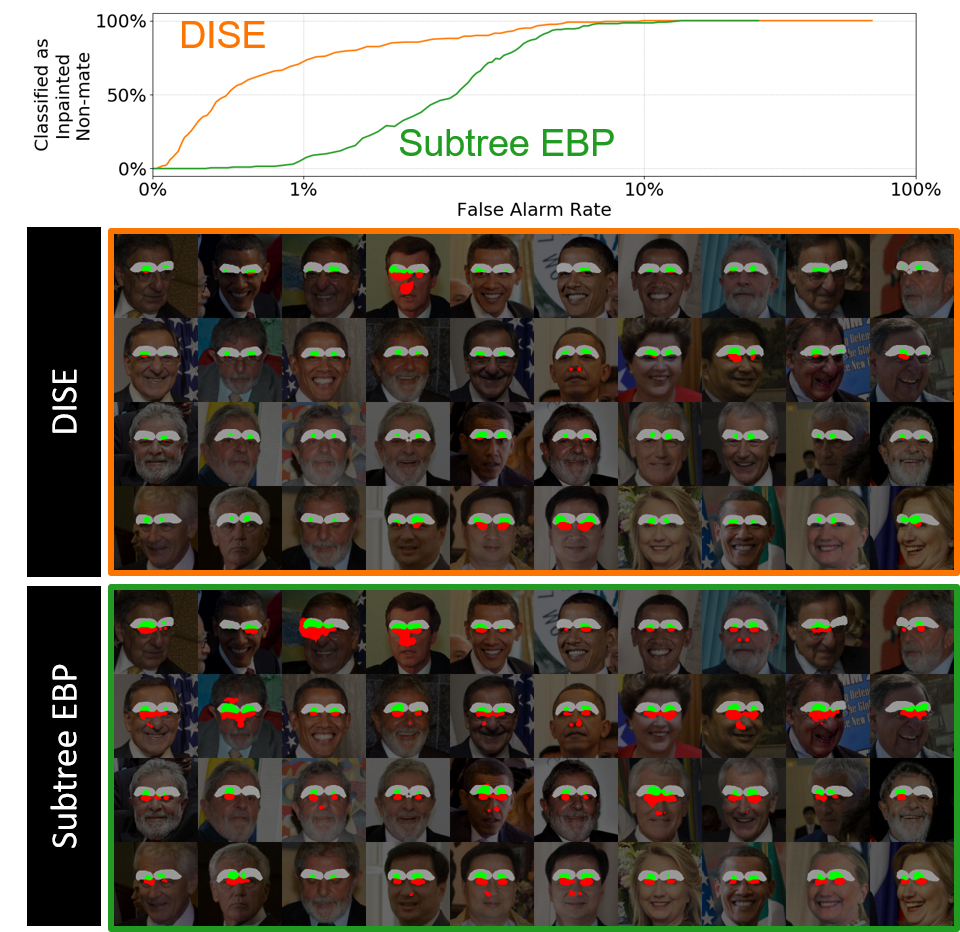}
    \caption{Eyebrow Mask (ResNet-101): Evaluation plot and classification on saliency maps from
    Subtree EBP  and
    DISE  at identity flip.}
\label{fig:ResNetv4_mask5} 
\end{figure*}

\begin{figure*}
    \centering
    \includegraphics[width=\linewidth]{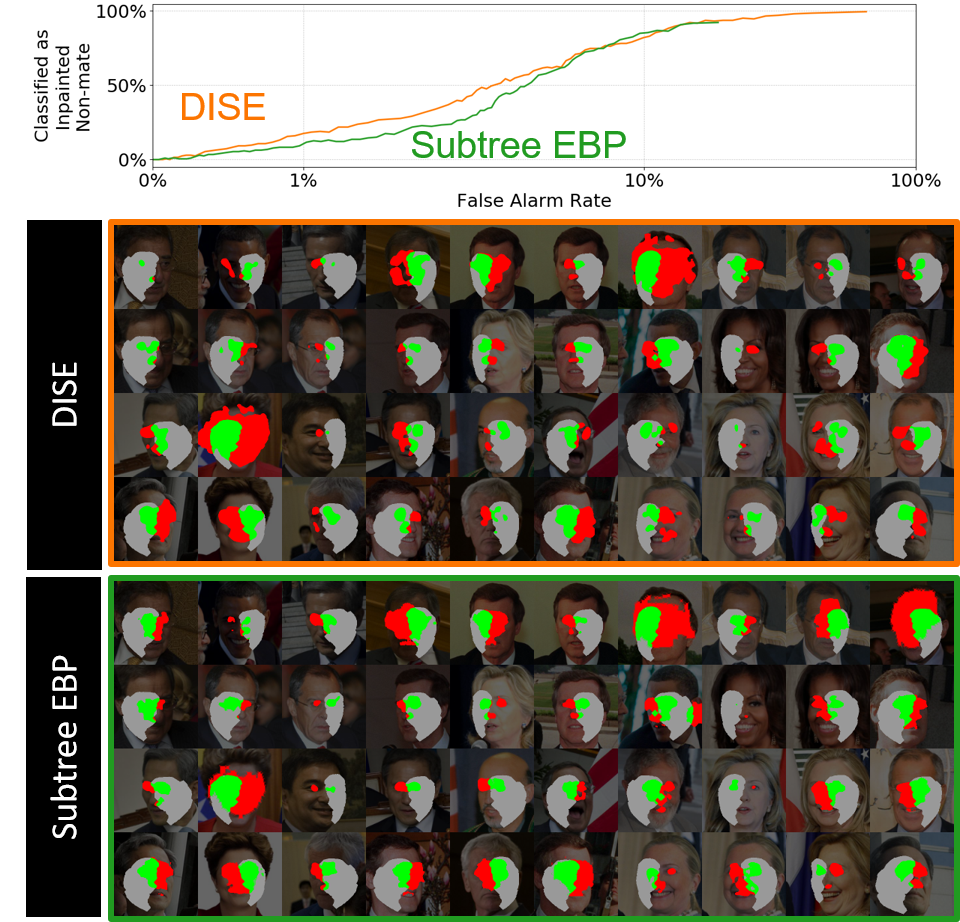}
    \caption{Left-/Right-Face Mask (ResNet-101): Evaluation plot and classification on saliency maps from
    Subtree EBP  and
    DISE  at identity flip.}
\label{fig:ResNetv4_mask167} 
\end{figure*}

\begin{figure*}
    \centering
    \includegraphics[width=\linewidth]{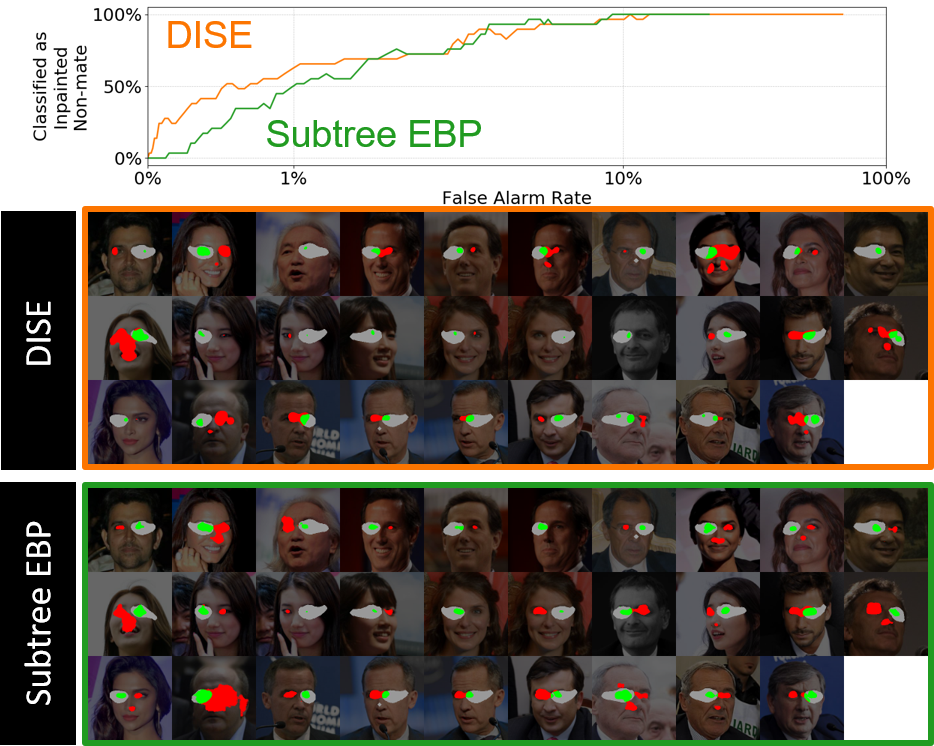}
    \caption{Left-/right- eye Mask (ResNet-101): Evaluation plot and classification on saliency maps from
    Subtree EBP  and
    DISE  at identity flip.}
\label{fig:ResNetv4_mask189} 
\end{figure*}

\begin{figure}[t]
    \centering
    \includegraphics[width=0.9\linewidth]{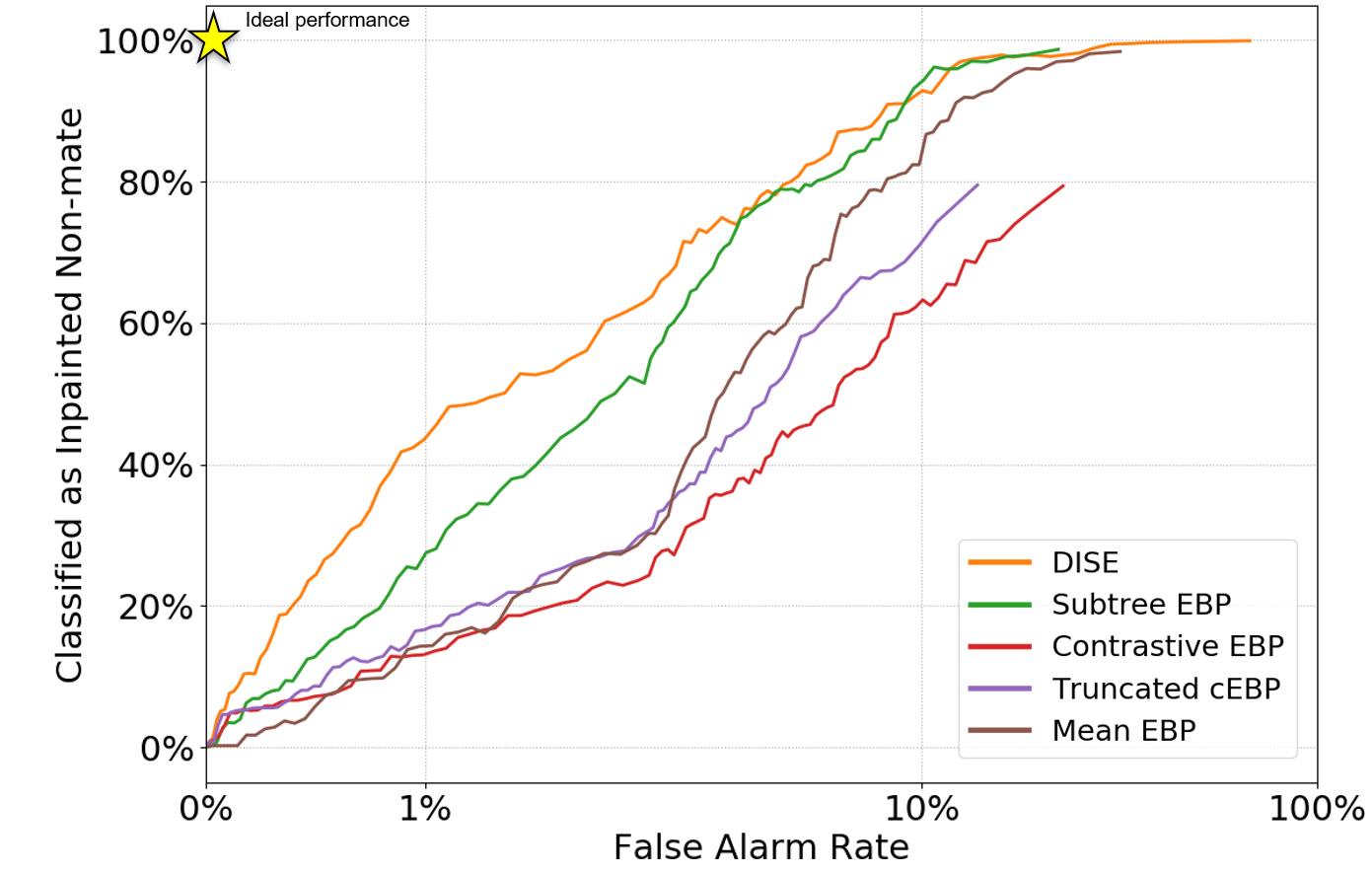}
    \caption{
    Inpainting game analysis using the ResNet-101.  
    }
    \label{f:inpainting_game_results_resnet101}
\end{figure}

\end{document}